\documentclass[pdflatex,sn-mathphys]{sn-jnl}

\usepackage{subfig}
\usepackage{comment}
\usepackage{soul}

\usepackage{makecell}

\usepackage[font=small,labelfont=bf]{caption}
\usepackage{array,ragged2e}

\usepackage{threeparttable}  

\usepackage[numbers]{natbib}

\newcommand{\tcolB}{\textcolor{blue}}

\newcommand{\tcolR}{\textcolor{red}}

\newcolumntype{C}[1]{>{\centering\arraybackslash}p{#1}}
\newcolumntype{P}[1]{>{\raggedright\arraybackslash}p{#1}}
\newcommand{\ignore}[1]{}
			
\algnewcommand{\LineCommentt}[1]{\State \(\triangleright\) #1}
\newcommand{\ABV}{ABV}
\newcommand{\ABVS}{ABVs}

\newcommand{\STAB}[1]{\begin{tabular}{@{}c@{}}#1\end{tabular}}
\begin{document}

\title[Token Classification for Disambiguating Medical Abbreviations]{Token Classification for Disambiguating Medical Abbreviations}

\author*[1]{\fnm{Mucahit} \sur{Cevik}}\email{mcevik@ryerson.ca}
\author[1]{\fnm{Sanaz} \sur{Mohammad Jafari}}
\author[1]{\fnm{Mitchell} \sur{Myers}} 
\author[1]{\fnm{Savas} \sur{Yildirim}} 

\affil[1]{
\orgname{Ryerson University}, \orgaddress{\street{44 Gerrard St E}, \city{Toronto}, \postcode{M5B 1G3}, \state{Ontario}, \country{Canada}}}

\abstract{
Abbreviations are unavoidable yet critical parts of the medical text.
Using abbreviations, especially in clinical patient notes, can save time and space, protect sensitive information, and help avoid repetitions.
However, most abbreviations might have multiple senses, and the lack of a standardized mapping system makes disambiguating abbreviations a difficult and time-consuming task.
The main objective of this study is to examine the feasibility of token classification methods for medical abbreviation disambiguation.
Specifically, we explore the capability of token classification methods to deal with multiple unique abbreviations in a single text.
We use two public datasets to compare and contrast the performance of several transformer models pre-trained on different scientific and medical corpora.
Our proposed token classification approach outperforms the more commonly used text classification models for the abbreviation disambiguation task.
In particular, the SciBERT model shows a strong performance for both token and text classification tasks over the two considered datasets.
Furthermore, we find that abbreviation disambiguation performance for the text classification models becomes comparable to that of token classification only when postprocessing is applied to their predictions, which involves filtering possible labels for an abbreviation based on the training data.
}

\maketitle
\section{Introduction} \label{sec:intro}
Word Sense Disambiguation (WSD) is the task of identifying the correct sense of an ambiguous word with several possible meanings given its context \citep{navigli2009word}.
For example, the term \textit{well} can be associated with health, i.e., \textit{I am feeling well} or a source of water, i.e., \textit{You can drink from the well}.
Humans have the innate ability to do this and often differentiate between word senses subconsciously.
However, automated WSD is considered to be one of the most challenging tasks in Natural Language Processing (NLP) \citep{agirre2007word}.
Abbreviation Disambiguation (AD) is a sub-task of WSD that focuses on accurately expanding or decoding ambiguous abbreviations (\ABV{}) in text data (e.g., ABI for \textit{Acquired Brain Injury}).
While humans are able to interpret common \ABVS{} to a degree, full-word WSD is a much more difficult task as they are usually domain-specific and require prior knowledge.
In this paper, we focus on AD in medical texts and, in this case, disambiguation can be in the form of providing the actual long-form version of the \ABV{} or outputting key additional context words associated with it so that any confusion about the sense is removed, assuming the reader is a medical professional.
Table~\ref{table: AD_outputs} provides examples of these two forms of AD.
\setlength{\tabcolsep}{7pt}
\renewcommand{\arraystretch}{1.15}
\begin{table}[!ht]
    \caption{Examples of AD outputs}
    \label{table: AD_outputs}
    \centering
    \resizebox{0.895\textwidth}{!}{
    \begin{tabular}{l l l l } 
     \toprule
\textbf{Output Type} & \textbf{\ABV{}} & \textbf{Label for Prediction Task} & \textbf{Actual \ABV{} Sense} \\ 
     \midrule
     Long Form Version & MBF & myocardial blood flow & Myocardial Blood Flow \\
     Key Context Word & IF & staining & Immunofluorescence Staining\\ 
     \bottomrule
    \end{tabular}
    }
\end{table}

In the medical domain, especially in clinical notes, abbreviations are extremely common due to their ability to increase efficiency and protect patients' privacy \citep{merriam-webster,jaber2022disambiguating}. One study found that an estimated 30-50\% of clinical notes were made up of abbreviations \citep{grossman2018method}. However, while there are benefits to \ABVS{}, they can sometimes be obscure to their actual sense making them un-intuitive and more difficult to decipher without additional information. Furthermore, \ABVS{} can have multiple meanings (long-form versions), and the actual sense depends heavily on the context. Moreover, new \ABVS{} are regularly being created without a standardized mapping system. This makes AD a difficult and time-consuming task that can delay the extremely important flow of communication in medical operations. These challenges in AD point to the necessity for building reliable alternatives using WSD techniques.

WSD is a heavily researched area in NLP and the majority of proposed methods can be grouped into three categories: knowledge-based~\citep{mcinnes2011using}, unsupervised~\citep{xu2012combining}, and supervised~\citep{devlin2018bert}. Knowledge-based methods rely mainly on dictionaries, sense inventories, and hand-crafted rules to predict the correct sense of an ambiguous word. Unsupervised methods do not need sense inventories and instead, mainly employ clustering methods to differentiate between different senses and contexts. Finally, supervised methods constitute the most commonly employed techniques for WSD; they typically require annotated data and classifiers trained over this data are used to detect the correct sense of the abbreviation. Common supervised WSD methods include Decision Trees (DT)~\cite{pakhomov2005abbreviation}, Support Vector Machines (SVM)~\citep{joshi2006comparative, moon2012automated}, Naive Bayes (NB)~\citep{jaber2021disambiguating}, and neural networks \citep{lee2020biobert}. 

Most research in AD looks at the task through text classification, where, given a piece of text with at least one \ABV{} present, a label is assigned to the whole text that decodes the \ABV{}. Multiple \ABVS{} in a text can complicate the text classification approaches, as it is possible to confuse the labels that are associated with the \ABVS{}. Therefore, in this study, we propose to frame the WSD problem as a token classification task, which can automatically handle multiple \ABVS{} occurring in the same text. We employ recent Named Entity Recognition (NER) methods to find the most effective token classification strategy. 
Well-known NER methods include probabilistic Conditional Random Fields (CRF) \citep{mccallum2012efficiently} and transformer models such as BERT~\citep{devlin2018bert}.

We conduct a detailed empirical analysis to investigate the feasibility of token classification methods for the WSD task with the objective of providing a more practical means to automatically expand or disambiguate medical \ABVS{} in text. We particularly focus on pre-trained BERT models to compare text and token classification for the WSD task. 
We compare these models against a CRF-based approach as well.
The main contributions of our study can be summarized as follows:
\begin{itemize}\setlength\itemsep{0.3em}
    \item To the best of our knowledge, this is the first study that considers state-of-the-art token classification methods for medical AD tasks.
    Our findings suggest that the token classification methods can outperform the text classification approaches for these tasks.
    
    \item We propose a postprocessing strategy for \ABV{} prediction, which filters the candidate set of labels for any individual \ABV{} based on the corresponding occurrences in the training data. We find that text classification methods benefit significantly from postprocessing. On the other hand, token classification methods, in part thanks to their ability to examine the data at a more granular level, do not require postprocessing to achieve a high-performance level.
    
    \item We conduct an extensive numerical study with two unique medical AD datasets and several text classification and token classification models.
    In this regard, our study contributes to a better understanding of the capabilities of these state-of-the-art methods for medical AD tasks.
    
    
\end{itemize}

The rest of the paper is structured as follows.
Section~\ref{sec:lit_review} explores previous literature on NER as well as medical AD and WSD tasks.
Section~\ref{sec:Methodology} provides an exploratory analysis of our datasets, and details the proposed methodologies, experimental design, and evaluation metrics.
Section~\ref{sec:Results} presents results from our detailed numerical study.
Finally, Section~\ref{sec:Conclusions} concludes the paper with a summary of our main findings and a discussion on future research directions.

\section{Related work} \label{sec:lit_review}
We briefly review the previous works on medical AD and NER and discuss the performance of different models and techniques in these tasks. AD and WSD research has become fairly popular in recent years and while they have been applied to several domains, we focus our review on WSD in the medical domain. A summary of the most closely related studies to our work is provided in Table~\ref{tbl:lit_rev_summary}.
\setlength{\tabcolsep}{4.5pt}
\renewcommand{\arraystretch}{1.05}
\begin{table}[!ht]
\caption{Summary of the relevant papers}
\label{tbl:lit_rev_summary}
\resizebox{0.99\textwidth}{!}{
\begin{tabular}{P{0.15\textwidth} P{0.24\textwidth} P{0.37\textwidth} P{0.34\textwidth}}
\toprule
{\bf Study} & {\bf Model} & {\bf Methodology} & {\bf Datasets}\\
\midrule
\citet{pakhomov2005abbreviation} & C5.0 Decision tree~\citep{quinlan2004data} & Leverage additional context found online to disambiguate \ABV{} & Clinical notes of Mayo Clinic$^1$\\
\midrule
\citet{xu2012combining} & Profile-based method & Combine sense frequency information with a profile-based method & NYPH$^2$ discharge summary
corpus and physician-typed hospital admission notes\\
\midrule
\citet{joshi2006comparative} & NB, DT and SVM & Include POS tags, unigrams and bigrams to improve the accuracy & Clinical notes of Mayo Clinic$^1$\\
\midrule
\citet{moon2012automated} & NB, DT and SVM & Optimize the window size for each orientation and determine the minimum training sample size & Clinical notes of Fairview Health Services$^3$\\
\midrule
\citet{wu2015clinical} & SVM & Investigate three different embedding methods to improve the feature set of SVM & Annotated \ABV{} of the Vanderbilt University Hospital’s admission notes and clinical notes of the University of Minnesota (UMN)\\
\midrule
\citet{jaber2021disambiguating} & NB, SVM & Investigate four strategies to use pre-trained word embedding as features & Clinical notes of the UMN\\
\midrule
\citet{li2019neural} & Traditional ML models, NN models and proposed ELMo+Topic model & Incorporate topic attention on ELMo word representation  & UMN, PubMed$^4$ and MIMIC-III~\citep{johnson2016mimic} \\ 
\midrule
\citet{jin2019deep} & DECBAE & Utilize BioELMo to extract features and pass them to BiLSTM & PubMed\\ 
\midrule
\citet{lee2020biobert}& BioBERT & Pre-train BERT on medical corpora & NCBI Disease~\citep{dougan2014ncbi}, GAD~\citep{bravo2015extraction}, BioASQ 6b-factoid~\citep{tsatsaronis2015overview}\\ 
\midrule
\citet{jaber2022disambiguating}& One-fits all classifier& Added simple neural network structure on top of the pre-trained BERT structure & UMN\\ 
\midrule
Our study & Adaptation of various BERT models & Incorporate token classification for AD task & MeDAL~\citep{wen2020medal} and UMN\\
\bottomrule
\end{tabular}
}
\\
{\tiny $^1$: from \url{https://www.mayoclinic.org/}}\\
{\tiny $^2$: from \url{https://www.nyp.org/}}\\
{\tiny $^3$: from \url{https://www.fairview.org/}}\\
{\tiny $^4$: from \url{https://pubmed.ncbi.nlm.nih.gov/}}\\

\end{table}

\citet{pakhomov2005abbreviation} were among the first to investigate medical acronym disambiguation and looked at the feasibility of semi-supervised learning for this task. By generating training data for each sense based on their context found across three large external sources including the World Wide Web, they were able to show the utility of leveraging massive publicly available data for medical AD. \citet{xu2012combining} built on this work and they not only included more external data sources for training data generation but also incorporated sense frequency information as an additional feature. 
While semi-supervised learning methods have shown encouraging performance when there is a lack of annotated corpora, supervised learning is a much more popular method for AD. \citet{joshi2006comparative} and \citet{moon2012automated} compared different supervised learning approaches, including NB, SVM, and DT for the WSD task. \citet{joshi2006comparative} extracted several key features from the AD datasets and found that all three models (NB, SVM, DT) performed fairly equally but overall, performance was maximized when all features were included in the training. \citet{moon2012automated} focused on minimizing training time and used these models to examine the impact of different window sizes around the target \ABV{} as well as finding the minimum training samples required for each label to achieve reasonable performance. 

The use of word embeddings for model training has become increasingly common in many NLP tasks, including text classification, information retrieval, and language translation. \citet{wu2015clinical} explored the impact of neural word embeddings trained on an extremely large medical corpus in the medical WSD task. On top of the traditional WSD feature-set, by adding two unique embedding-based features to their SVM classifier, one that took the max score of each embedding dimension from all surrounding words and another one that took the sum of the embedding vectors from the surrounding words, their model was able to achieve state-of-the-art results on their test datasets \citep{wu2015clinical}.  \citet{jaber2021disambiguating} conducted a similar study and compared different word embedding strategies on two different models: SVM and NB. Their results showed that SVM outperformed NB overall, and the best performance was achieved when using word embeddings generated from both medical and general data sources.

Deep learning-based approaches have been popular for AD tasks in recent years. 
\citet{li2019neural} proposed a neural topic attention model for the medical AD task where they took a few-shot-learning approach that combined topic attention and contextualized word embeddings learned from ELMo \citep{peters2018}. Applying the topic information and word embeddings to a Long-Short-Term-Memory (LSTM) model yielded the best results in their experiments. \citet{jin2019deep} proposed the Deep Contextualized Biomedical \ABV{} Expansion (DECBAE) model that utilizes BioELMo \citep{jin2019probing} word embeddings, a domain-specific version of ELMo \citep{peters2018}, and a fine-tuned Bidirectional-LSTM (BiLSTM) model to achieve state-of-the-art performance. Similar to ELMo and its descendent BioELMo, the pre-trained BERT\textsubscript{base} \citep{devlin2018bert} models offer complex word embeddings that apply to a wide array of topics. However, minimal or no exposure to key biomedical terms limits the benefits of transfer learning in this domain. Accordingly, \citet{lee2020biobert} proposed BioBERT, a pre-trained model for biomedical text mining.
\citet{jaber2022disambiguating} explored different BERT variants for the medical AD task and found that the variations pre-trained on medical text such as BioBERT outperformed BERT\textsubscript{base}. 

While many studies on medical AD referenced above consider the same objective as in our study, they mainly focus on single- or multi-label classification. We set out to investigate the potential of NER methods for the WSD task. NER is a token classification technique that identifies and categorizes key information and entities in unstructured text. There are three traditional approaches to address the NER problem: rule-based \citep{hanisch2005prominer,quimbaya2016named}, unsupervised learning \citep{zhang2013unsupervised}, and feature-based supervised learning \citep{settles2004biomedical,yao2015biomedical}.
The majority of NER studies follow the supervised learning approach and accordingly, several classifiers have been explored for NER over the years. However, the CRF model has been shown to be highly effective for the NER tasks due to its ability to capture the context around the target word or entity. \citet{mccallum2012efficiently} proposed a feature induction method for CRF in NER. 
By iteratively adding features and only focusing on the ones that maximize the log-likelihood, they were able to greatly improve the NER performance over a fixed feature set approach, while also controlling the training time. 
More recently, transfer learning has become popular for NER with the advent of ELMo and BERT models \citep{devlin2018bert, peters2018}. 
\citet{souza2019portuguese} proposed a BERT-CRF model in a Portuguese NER task. Additionally, Biomedical Named Entity Recognition (BioNER) was developed to identify entities such as genes, proteins, diseases, chemicals, and species, in medical texts and clinical notes. 
For instance, \citet{liu2020k} proposed K-BERT, a BERT model that can be injected with domain-specific knowledge more efficiently than pre-training. 
In several domain-specific NER tasks, including a medical NER task, K-BERT was shown to outperform regular BERT. 

The methods discussed above are highly relevant to our study, particularly the work by \citet{jaber2022disambiguating}. 
Therefore, these relevant techniques are implemented as baselines in our numerical study. 
We note that the key difference between our work and the previous literature is that we apply NER methods for a medical WSD task. 
Not only does our model have to predict whether a token is an \ABV{}, but it also needs to output the correct label depending on the context. 
While this is similar to regular BioNER, in BioNER there are usually only a few (e.g., $<10$) possible NER labels, whereas, in AD/WSD tasks, we are required to deal with more than 1,000 labels.

\section{Methodology} \label{sec:Methodology}
In this section, the token classification methods employed in our analysis are explained and the structure of the classification models and datasets are reviewed. 
Furthermore, the details of the experimental setup, hyperparameter tuning and evaluation metrics are provided.

Recent studies on WSD mainly focus on text classification where, given a piece of text with an abbreviation present, the entire text is labeled with the correct sense of the \ABV{}. However, since datasets can have multiple unique abbreviations in each instance, we take an alternative approach to solving the WSD task. While multi-label classification methods can also be considered for this task, the challenge would then become to ensure that each predicted label is correctly assigned to its respective \ABV{}.  Further complications might also arise when the number of labels predicted does not match the number of target \ABVS{}. Token classification, on the other hand, allows us to assign a single label to each token in the text and therefore, it enables assigning each abbreviation to its corresponding sense.
As such, it allows easy interpretation of the predictions and adoption of these methods beyond these experiments.

\subsection{Classification models}
We provide the details of each of the seven models used in our experiments below. CRF and BiLSTM are considered as baseline models and we employ five BERT variants for our main comparative analysis: DistilBERT, BioBERT, BlueBERT, MS-BERT, and SciBERT. 
    \subsubsection{CRF}
    CRF is a model that uses a probabilistic approach in modeling sequential data \citep{lafferty2001conditional}. CRFs are popular in part-of-speech (POS) tagging, NER, and other token classification tasks due to their ability to learn sequential contexts in text and utilize domain and data-specific handcrafted features to predict the label for each token. For our experiments, several features are designed for the CRF model including word components such as prefix and suffix, capitalization, and a check flag to determine whether each token is an abbreviation or not. We also captured contextual features that examined the nearest neighbor on either side of the current word (i.e., with window size $[-1,1]$). While this model is commonly used for sequence labeling tasks, one drawback is its computational complexity as training time drastically increases with the sequence length and the number of labels. Accordingly, we consider the CRF model as a baseline for our experiments on a reduced-label dataset, which contains only a subset of all the available labels.

    \subsubsection{BiLSTM}
    An LSTM model is a recurrent neural network (RNN) that can find and maintain long-range dependencies in data \citep{hochreiter1997long, jozefowicz2015empirical}. Due to the problem of vanishing gradients, regular RNNs are limited by how much important information can be stored in their \textit{memory}. However, with LSTMs, thanks to a complex gating system, key information no matter how far back in the sequence it is, can be maintained and accessed for the prediction task. A popular extension to the LSTM model is the bidirectional LSTM (BiLSTM), which has the ability to learn long-range dependencies in both directions -- from left to right and right to left. The BiLSTM model has become a common foundation for several sequence tagging tasks \citep{huang2015bidirectional} and it is considered as a baseline in our analysis.

    \subsubsection{BERT}
    BERT is a pre-trained language model and stands for Bidirectional Encoder Representations from Transformers \citep{devlin2018bert}. BERT's model architecture is based on the Transformer model proposed by \citet{vaswani2017attention}, which is an attention mechanism that can be used to learn relationships between words and sub-words. Similar to the BiLSTM, the bidirectional nature of this model allows learning the context of a word based on the words both from left and right. A notable challenge in pre-training is defining a prediction task. As the main training strategy, BERT employs a masked language model objective that randomly masks a proportion of tokens in the input. The task is to predict the actual masked word based on its context. Additionally, BERT employs a next sentence prediction task which requires the model to predict whether the second of two input sentences actually follows the first. In pre-training stage for these tasks, the BooksCorpus of 800M words and English Wikipedia of 2,500M words are employed. BERT has achieved state-of-the-art performance on several NLP tasks.
    Accordingly, five BERT variants are considered in our experiments.
    These include a popular lighter BERT variant (DistillBERT), three models that are pre-trained on different medical corpora (BioBERT, BlueBERT, MS-BERT) and another model that is pre-trained on scientific text with its custom domain-specific vocabulary (SciBERT).
    Below, we briefly summarize each of these BERT variants.
    
    \begin{itemize}\setlength\itemsep{0.3em}
    \item \emph{DistilBERT}: 
    This model is the \textit{faster, cheaper, and lighter} version of the original BERT~\citep{sanh2019distilbert}. While full-size transformers-based models offer outstanding performance, their usage is limited by high computational complexity. DistilBERT is based on the same architecture as BERT but in a condensed and more efficient form. DistilBERT was shown to retain 97\% of BERT's performance while being 60\% faster~\citep{sanh2019distilbert}. 
    
    \item \emph{BioBERT}:
    BioBERT is a BERT variant that is additionally pre-trained on large-scale biomedical corpora to excel in biomedical text-mining tasks.
    Specifically, it is pre-trained on PubMed abstracts that contain 4.5B words and PubMed Central full-text articles with 13.5B words.
    
    \item \emph{BlueBERT}:
    Similar to BioBERT, BlueBERT (Biomedical Language Understanding Evaluation) was pre-trained on biomedical data \citep{peng2019transfer}. This BERT variation trained on a very large corpus of more than 4B words from PubMed abstracts and over 500 million words from MIMIC-III clinical notes. 
    
    \item \emph{MS-BERT}:
    This model is an extension of BlueBERT that is further pre-trained on over 35 million words extracted from multiple sclerosis clinical notes collected between 2015 and 2019 in Toronto (hence the \textit{MS} in MS-BERT) \citep{NLP4H/msbert}. 
    
    \item \emph{SciBERT}:
    This model has two key differences from BERT\textsubscript{base}: its pre-training corpus and vocabulary~\citep{beltagy2019scibert}. Similar to the other domain-specific variations, SciBERT is pre-trained on scientific literature. The corpus consisted of 1.14 million articles from Semantic Scholar both from the computer science and biomedical domains, resulting in 3.17B words. 
    Unique only to SciBERT, however, is its vocabulary.
    Generating its own vocabulary instead of reusing the one from BERT\textsubscript{base} allowed SciBERT to capture the most frequently occurring words and sub-words from its specific domain.
    \end{itemize}

\subsection{Datasets}
We use two distinct medical text datasets in our experiments: MeDAL \citep{wen2020medal} and UMN \citep{moon_pakhomov_melton_2012}. The MeDAL dataset consists of medical abstracts where certain long-form words have been manually swapped with their abbreviated form. A key feature of this dataset is the occurrence of multiple unique \ABVS{} in one abstract. The UMN dataset, curated by the University of Minnesota's Digital Conservancy, is a collection of raw, anonymized clinical patient notes. There is only one target \ABV{} with an associated label that can occur multiple times in this dataset. The MeDAL dataset offers the unique challenge of dealing with multiple unique \ABVS{} in one instance and hence it provides a strong motivation for the use of token classification methods. On the other hand, the UMN dataset potentially provides a more likely real-life application of AD, since it is composed of raw clinical notes. The following section describes each dataset in detail.

\subsubsection{MeDAL dataset}
In our analysis, 2\% of the full 14 million abstract MeDAL dataset was used. In this subset, there are 288,080 rows and each row contains an abstract, the location of each \ABV{} given by its index, and the corresponding long-form versions or senses. An example of a MeDAL dataset instance can be found in Table~\ref{table: medal_example}. 
\setlength{\tabcolsep}{3pt}
\renewcommand{\arraystretch}{1.05}
\begin{table}[!ht]
\caption{A sample MeDAL data instance}
\label{table: medal_example}
\centering
\resizebox{0.859\textwidth}{!}{
\begin{tabular}{P{0.65\linewidth}  P{0.10\linewidth}  P{0.25\linewidth} } 
 \toprule
 \multicolumn{1}{l}{\textbf{Text}} & \multicolumn{1}{l}{\textbf{Locations}} & \multicolumn{1}{l}{\textbf{Labels}}\\ 
 \midrule
The kinetic disposition and betaadrenergic blocking action in relation to plasma level of a single oral dosey proportional to the extent of the histamine release it is concluded that the reduction in the in vitro amine uptake after anaphylactic and compound induced histamine release is due to the fact that there are a fewer intact granules capable of storing histamine and not primarily due to a damage to the mechanisms by which mast cells take up \textbf{\tcolB{BA}} in vitro the observations further strengthen the view that anaphylactic and compound induced \textbf{\tcolR{HR}} are noncytolytic processes & [76, 90] & [``\tcolB{Biogenic Amines}", ``\tcolR{Histamine Release}"] \\
 \bottomrule
\end{tabular}}
\end{table}

There are 557,248 unique words, 4,866 unique \ABVS{}, and 16,299 unique labels in the MeDAL dataset. However, to further reduce computational complexity and training time in our experiments, only the 300 most frequent \ABVS{} and their 1,005 most frequent labels were selected. This restriction reduced total rows to 147,728 and unique words to 320,168. 
 
\begin{figure}[!ht]
        \centering
      
        \subfloat[Word count distribution ]{\includegraphics[width=0.5\textwidth]{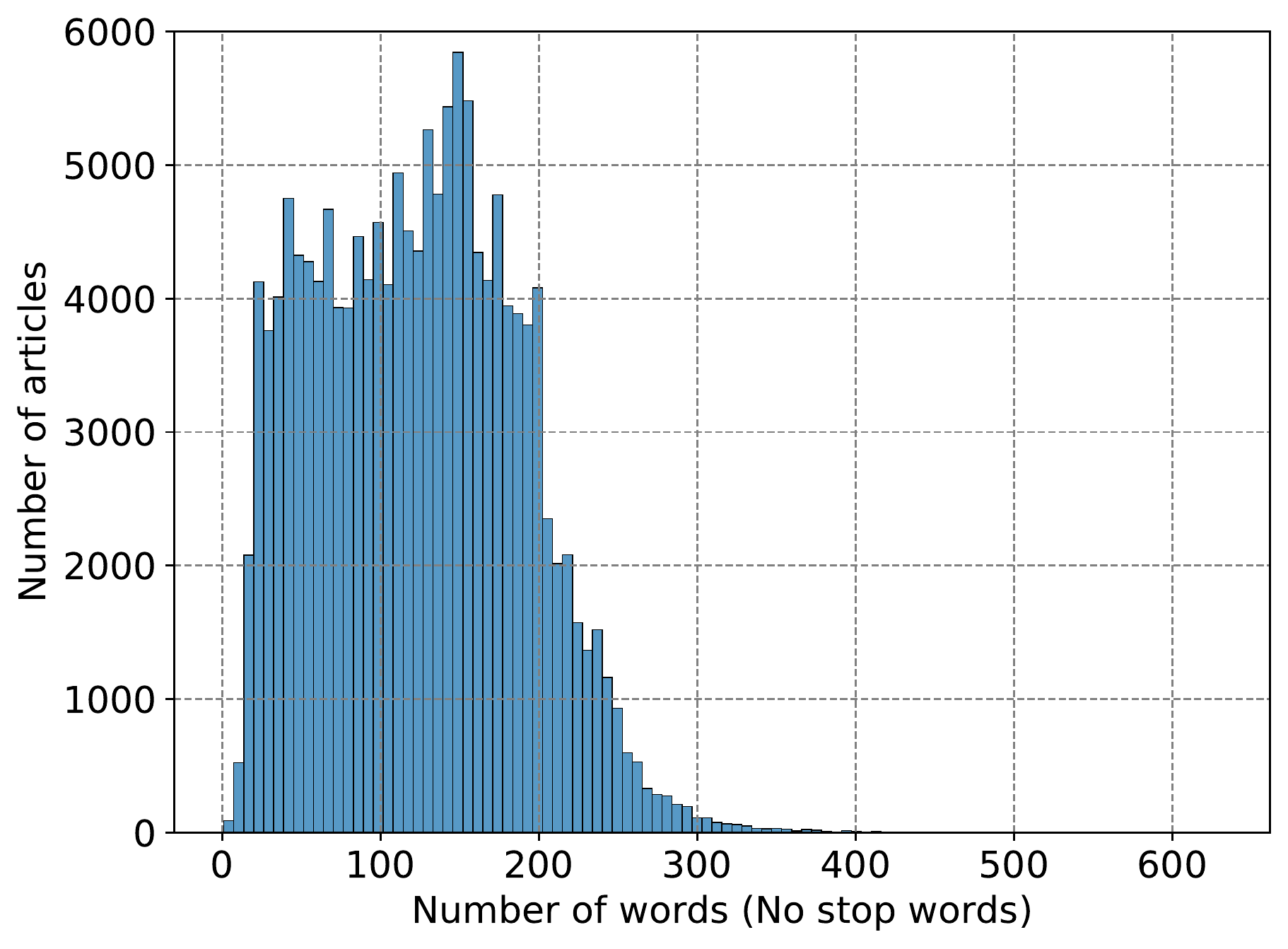}}
         \subfloat[Abbreviation count distribution ]{\includegraphics[width=0.5\textwidth]{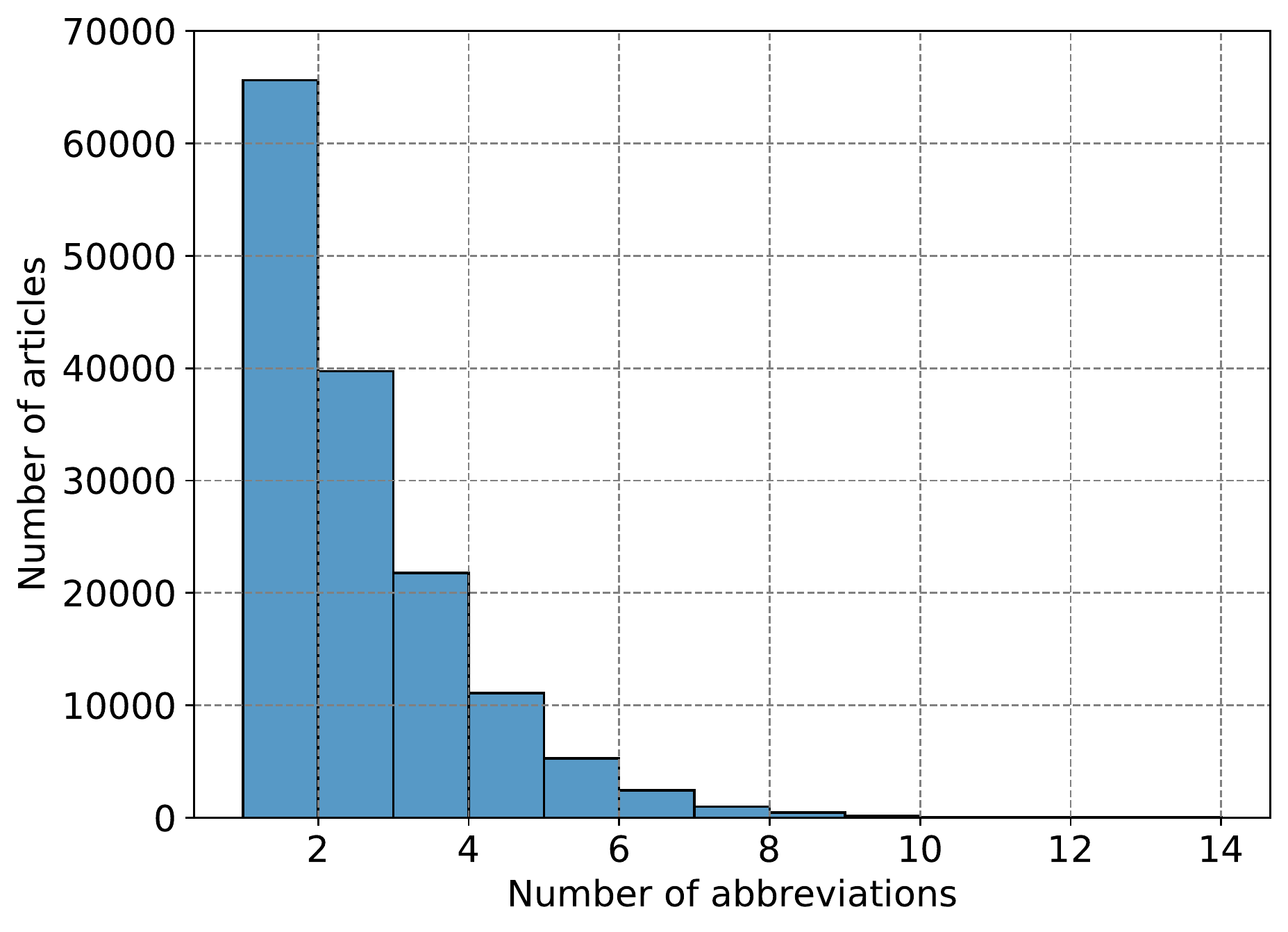}}
        \caption{Distribution of number of words and abbreviations for MeDAL dataset}
        \label{figure:MeDAL - \ABV{} and Word Count}
\end{figure}

\begin{figure}[!ht]
    \centering
    \subfloat[\label{fig:medal_abvmostsenses}Most ambiguous \ABVS{} (Greatest number of unique labels)]{\includegraphics[width=0.45\textwidth, height=4cm]{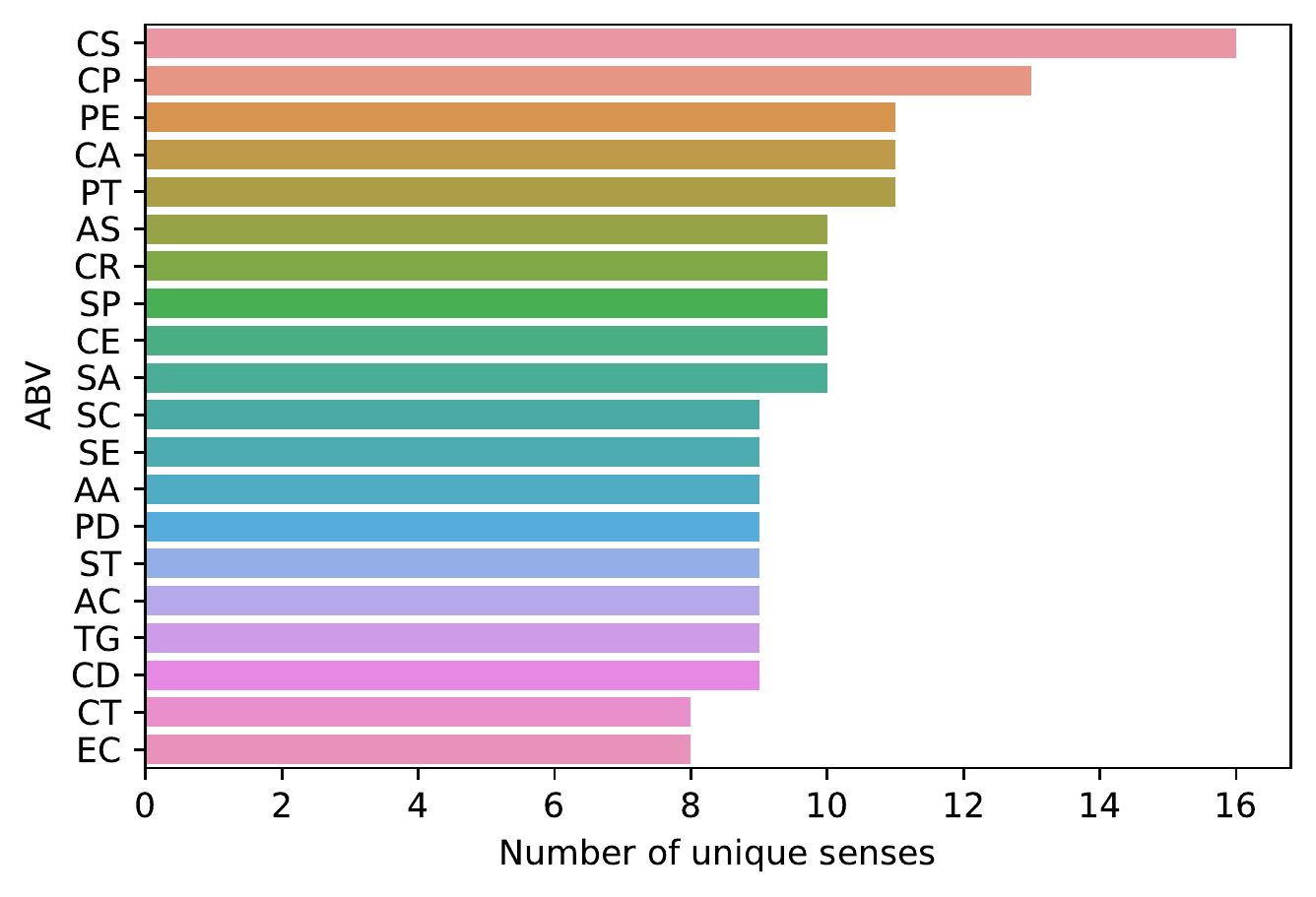}}
    \subfloat[\label{fig:medal_abvmostsenses_most_freq}Most frequent label of ambiguous \ABVS{} ]{\includegraphics[width=0.55\textwidth,height=4cm]{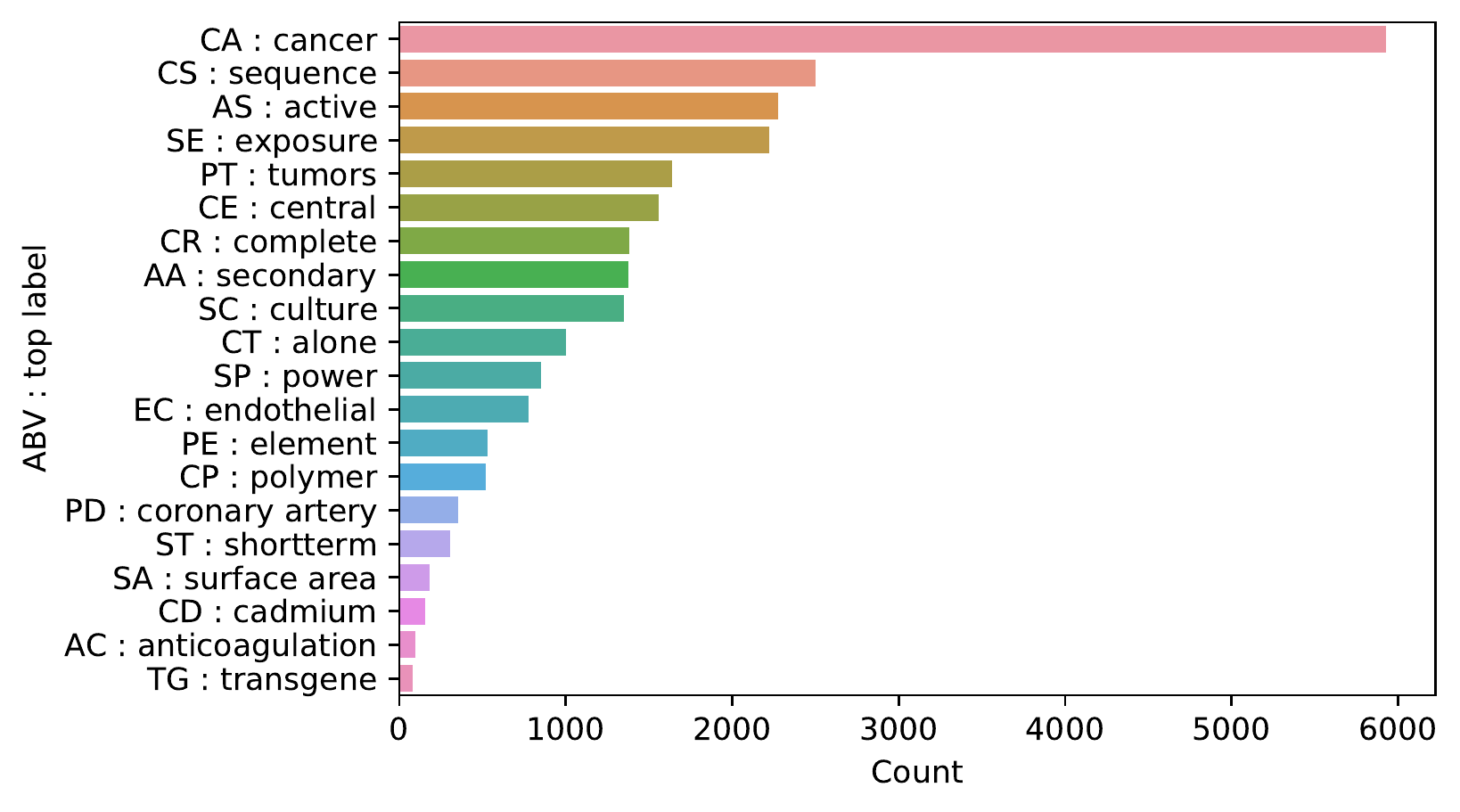}}
    \caption{Most ambiguous \ABVS{} and their most frequently occurring labels in the MeDAL dataset}
    \label{figure:MeDAL - ambiguous_ABVs}
\end{figure}

Figure~\ref{figure:MeDAL - \ABV{} and Word Count} illustrates \ABV{} and word count distribution across abstracts. The mean \ABV{} count is 2.1 while the mean word count is 124. However, by removing the stop words like \textit{the}, \textit{at}, or \textit{how}, the average word count drops to 77. In this subset, there are on average 3.35 unique senses associated with each abbreviation, with a minimum of 1 possible label (least ambiguous) to a maximum of 18 (most ambiguous). Figure~\ref{figure:MeDAL - ambiguous_ABVs} shows the 20 \ABVS{} with the greatest number of unique labels (see Figure~\ref{fig:medal_abvmostsenses}), which are considered the most ambiguous, as well as the label from each \ABV{} with the greatest number of occurrences in the dataset (see Figure~\ref{fig:medal_abvmostsenses_most_freq}). 
 
\begin{figure}[!ht]
    \centering
    \subfloat[\label{fig:MEDALbi_gram}Bi-grams distribution ]{\includegraphics[width=0.45\textwidth, height=4cm]{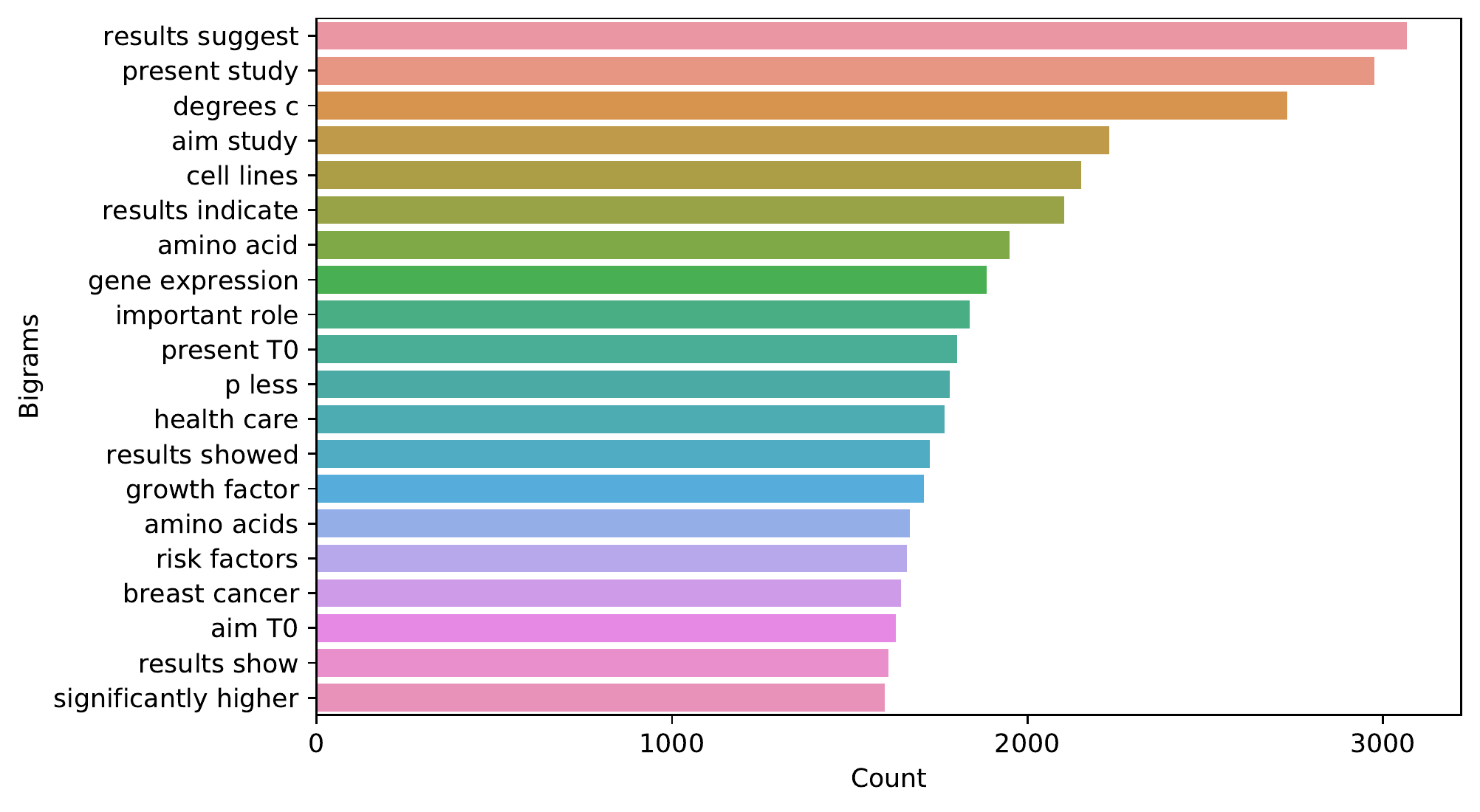}}
    \subfloat[\label{fig:MEDAL_tri_gram}Tri-grams distribution ]{\includegraphics[width=0.55\textwidth, height=4cm]{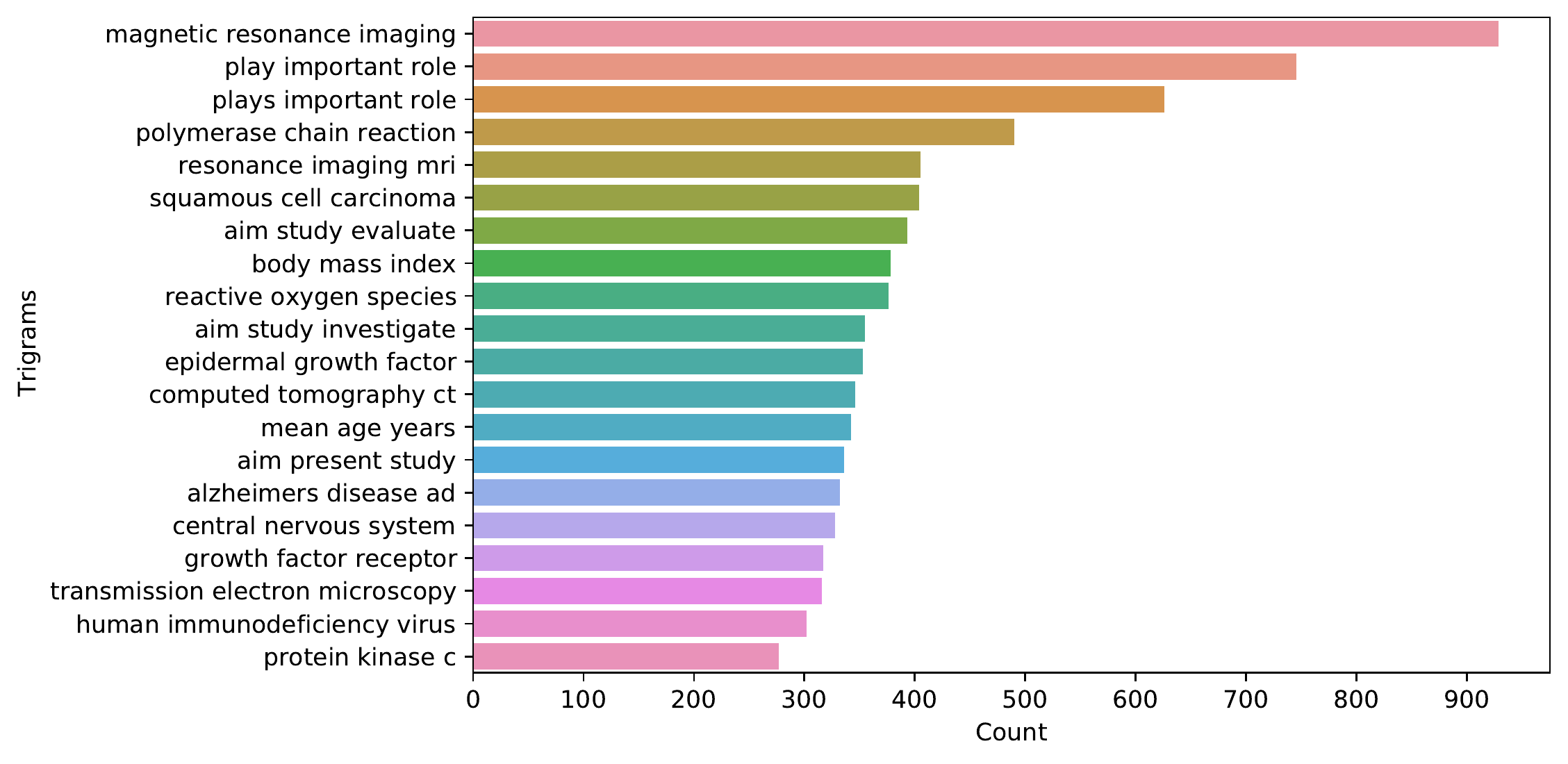}}
    \caption{Top 20 most frequent bi- and tri-grams in the MeDAL dataset}
    \label{figure:MeDAL - Bi and Tri-grams}
\end{figure}

These figures show that the dataset is imbalanced, specifically, the distribution of the number of examples across all MeDAL labels span from a minimum of 14 up to over 18,000. Figures~\ref{fig:MEDALbi_gram} and ~\ref{fig:MEDAL_tri_gram} display the most frequent bi- and tri-grams. The $n$-gram phrases and sample text in Table~\ref{table: medal_example} show that most of the text is well structured and most terms revolve around the themes of study, research, and medical experimentation.

\subsubsection{UMN dataset}
\begin{figure}[!ht]
    \centering
    \includegraphics[width = 0.6\textwidth]{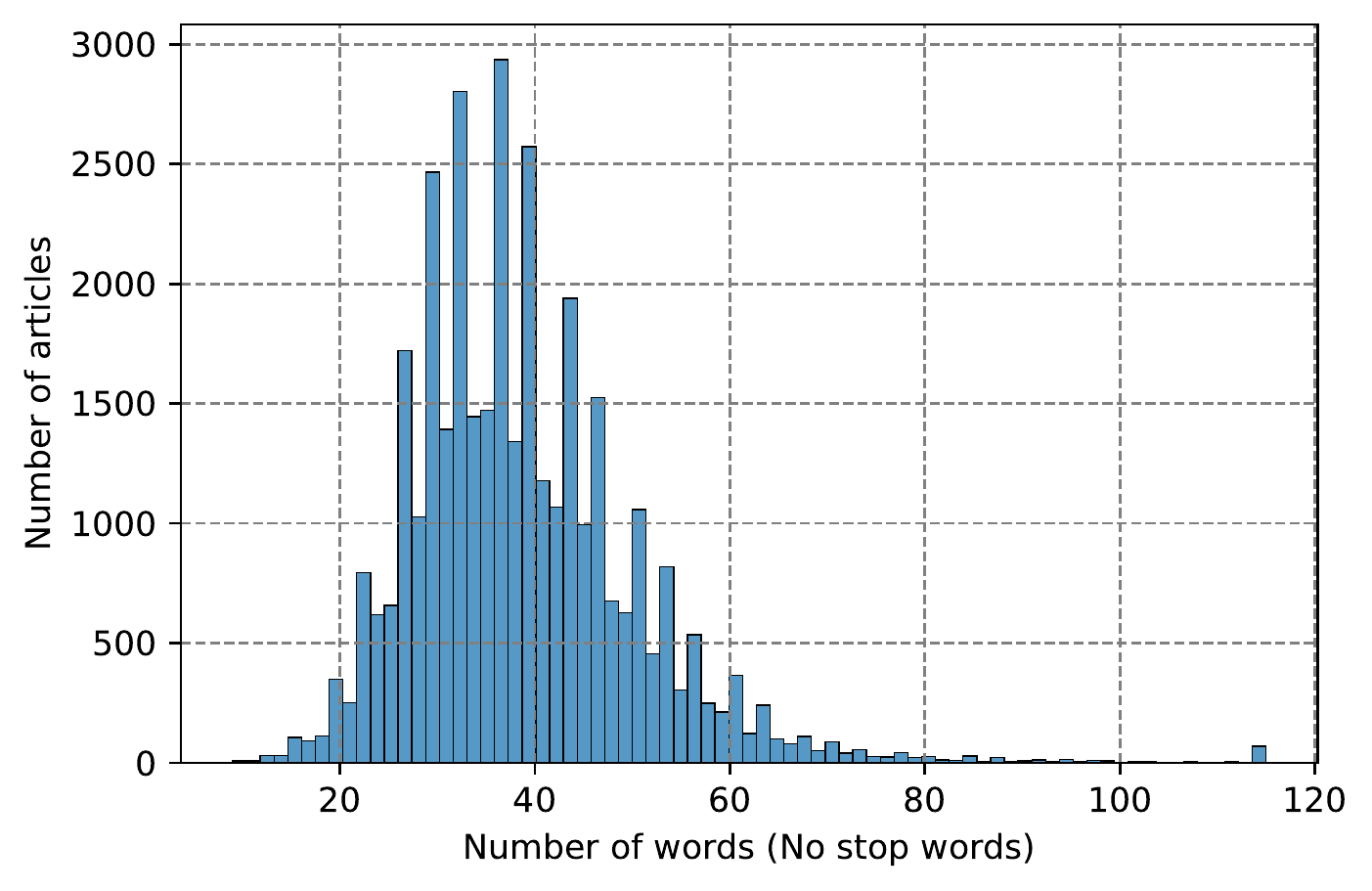}
    \caption{Word count distribution of the abstracts in the UMN dataset}
    \label{figure:umn_word_count}
\end{figure}

The UMN dataset consists of 36,996 rows, 74 unique \ABVS{}, and 346 unique labels. There are a total of 39,110 unique words in the text column. To reduce the complexity of the dataset, any label with fewer than five examples is dropped from the dataset, resulting in a final total of 203 labels and 72 \ABVS{}. Only including these labels in the dataset reduces total rows to 35,518 and total words to 37,283. As seen in Figure~\ref{figure:umn_word_count}, the word count distribution is centered around its mean of 39 words, or 59 including stop words. Table~\ref{table:UMN_example} presents a sample instance of the UMN dataset, which has a very similar format to the MeDAL dataset. However, in each row, there can only be one unique \ABV{} present that can occur multiple times.

  \setlength{\tabcolsep}{3pt}
\renewcommand{\arraystretch}{1.05}
 \begin{table}[!ht]
     \caption{A sample UMN data instance}
    \centering
    \resizebox{0.859\textwidth}{!}{
    \begin{tabular}{P{0.65\linewidth}  P{0.10\linewidth}  P{0.25\linewidth} } 
     \toprule
     \multicolumn{1}{c}{\textbf{Text}} & \multicolumn{1}{c}{\textbf{Locations}} & \multicolumn{1}{c}{\textbf{Labels}}\\ [0.5ex] 
     \midrule
    Her \textbf{\tcolR{PA}} pressures were 44/26 with a wedge of 22 with a CVP of 10 and her heart rate of 120 to 139. Her cardiac index was 3. Nursing made aggressive attempts to bring her \textbf{\tcolR{PA}} pressures down and on the day 2 of admission, she was found to have her Nipride running at 7 mcg/kg/minute. This was quickly weaned, and captopril instituted with hydralazine as needed. & [1, 35] & [``\tcolR{Pulmonary Artery}", ``\tcolR{Pulmonary Artery}"] \\
         
     \bottomrule
    \end{tabular}
    }
    \label{table:UMN_example}

\end{table}

\begin{figure}[!ht]
        \centering
        \subfloat[\label{fig:amb_UMN} Most ambiguous \ABVS{} (Greatest number of unique labels)]{\includegraphics[width=0.4\textwidth, height=4cm]{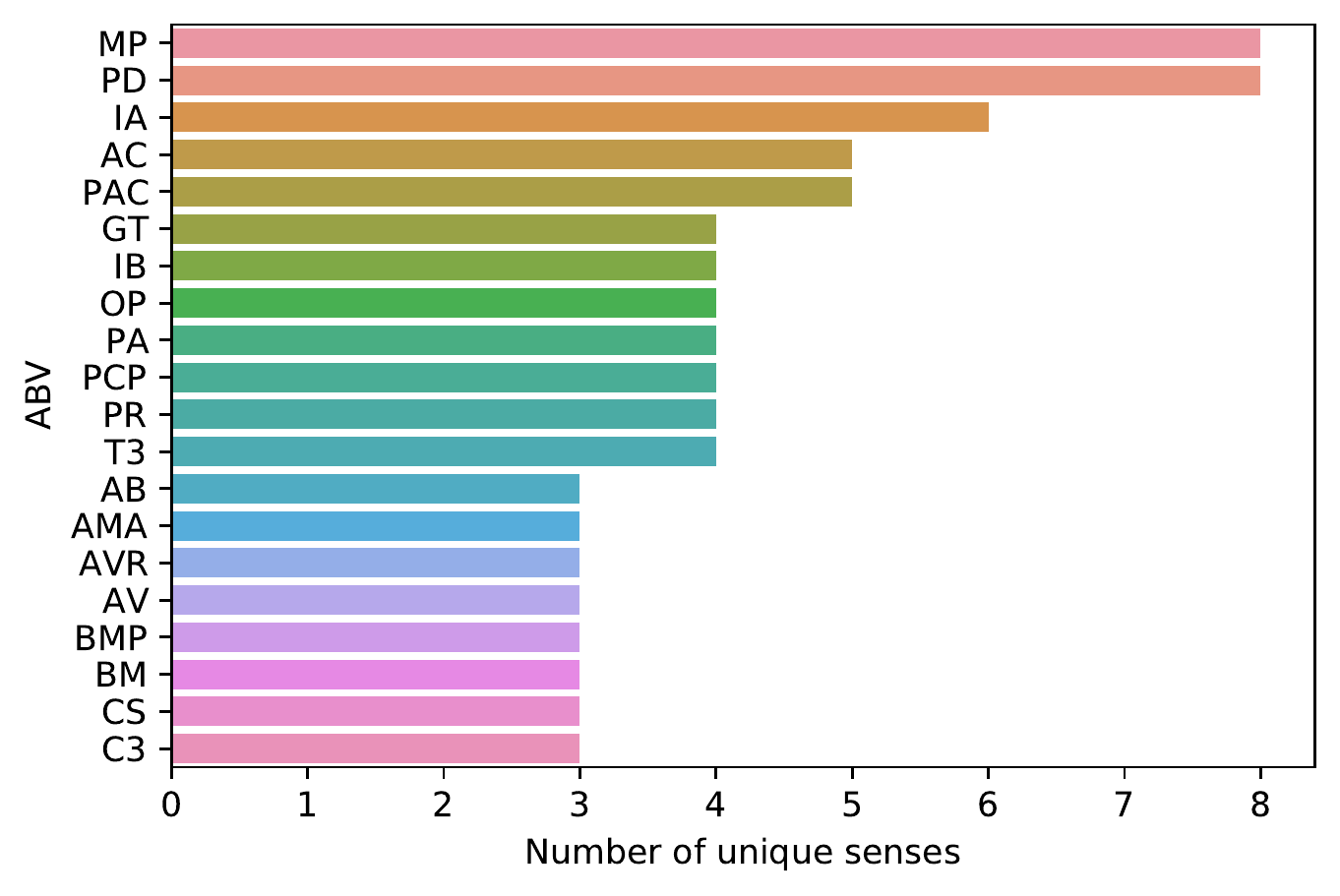}}
        \subfloat[\label{fig:label_UMN} Most frequent label of ambiguous \ABVS{}]{\includegraphics[width=0.6\textwidth, height=4cm]{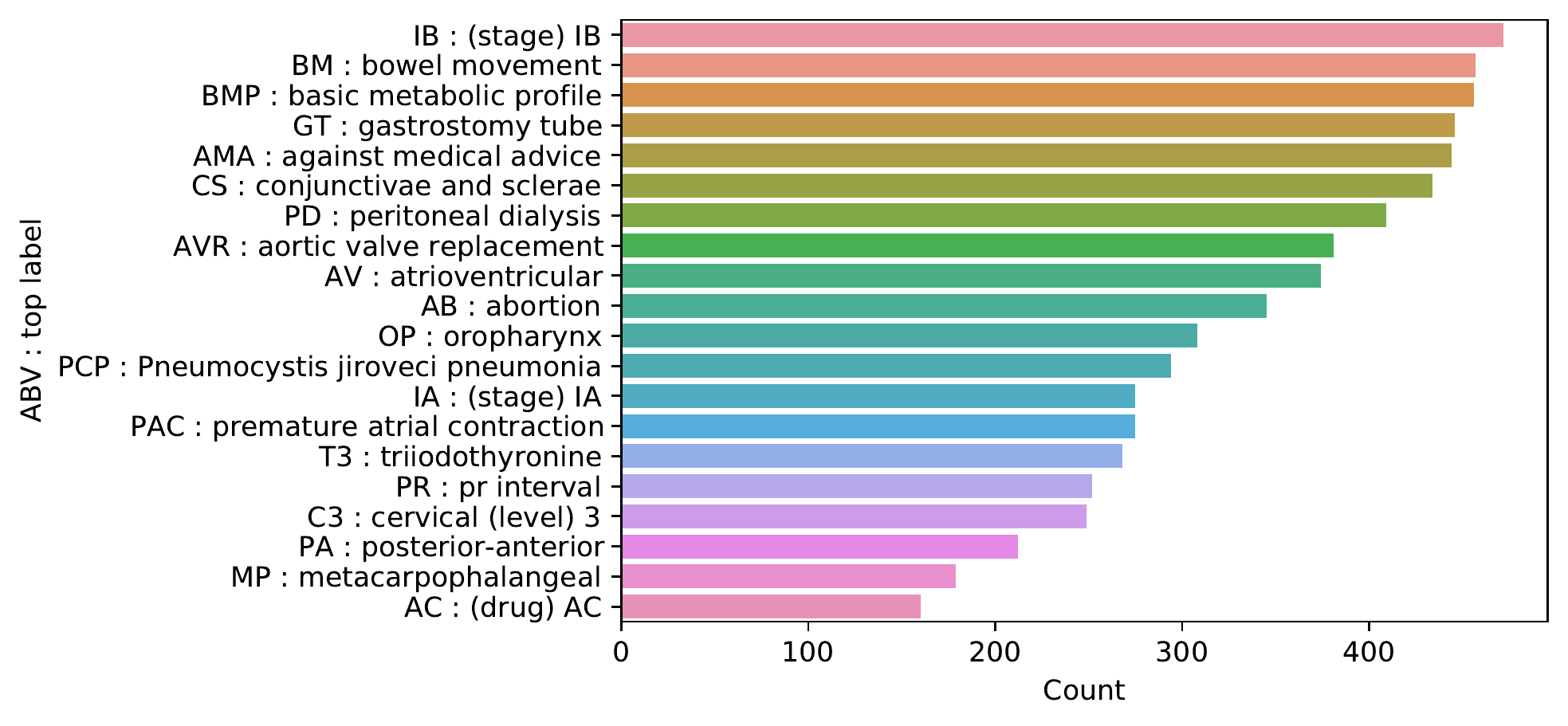}}
        \caption{Most ambiguous \ABVS{} and their most frequently occurring label in the UMN dataset}
        \label{figure:umn_ambiguous_abv}
\end{figure}

\begin{figure}[!ht]
        \centering
        \subfloat[\label{fig:bi_UMN} Bi-grams distribution ]{\includegraphics[width=0.5\textwidth, height=4cm]{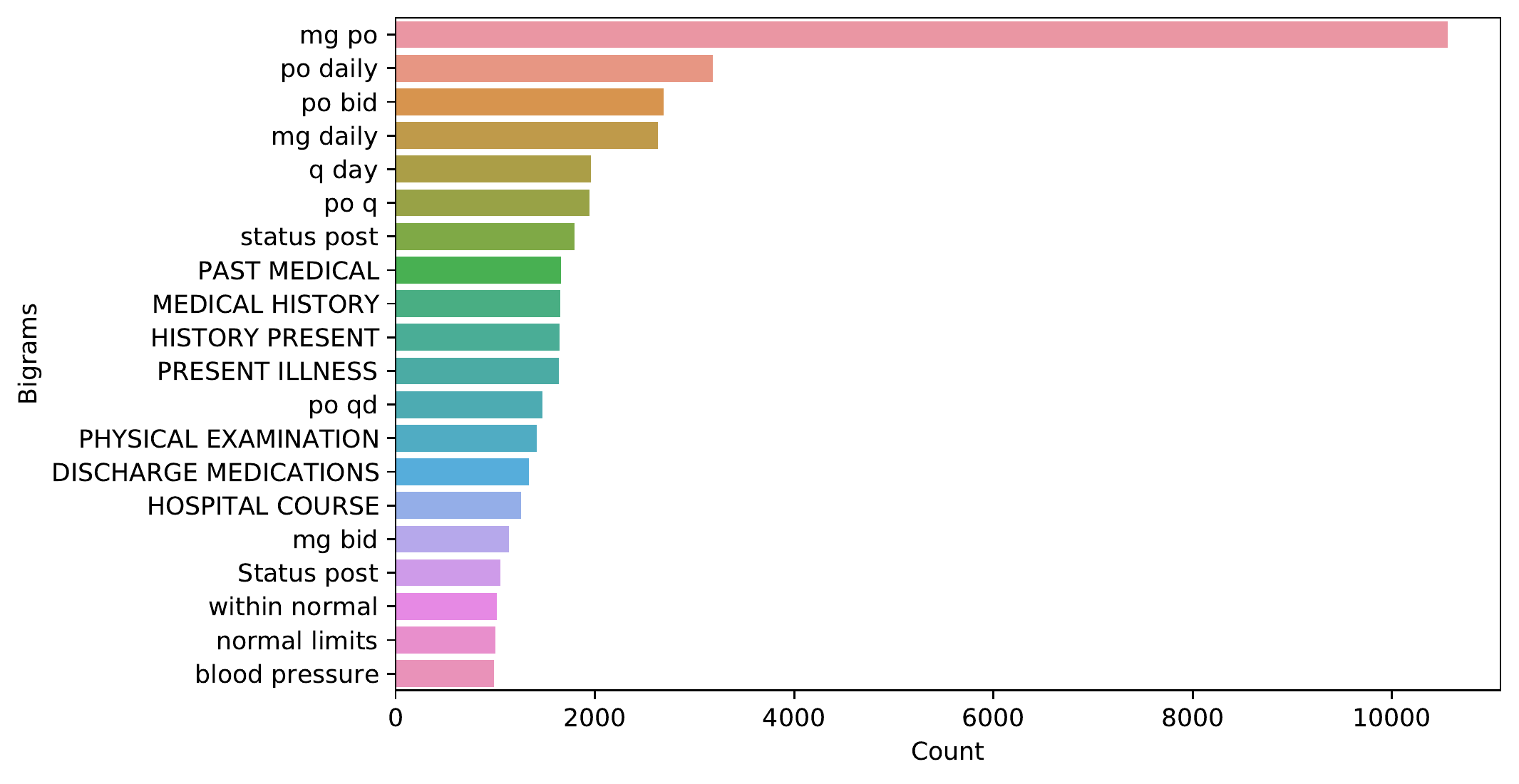}}
        \subfloat[\label{fig:tri_UMN} Tri-grams distribution ]{\includegraphics[width=0.5\textwidth, height=4cm]{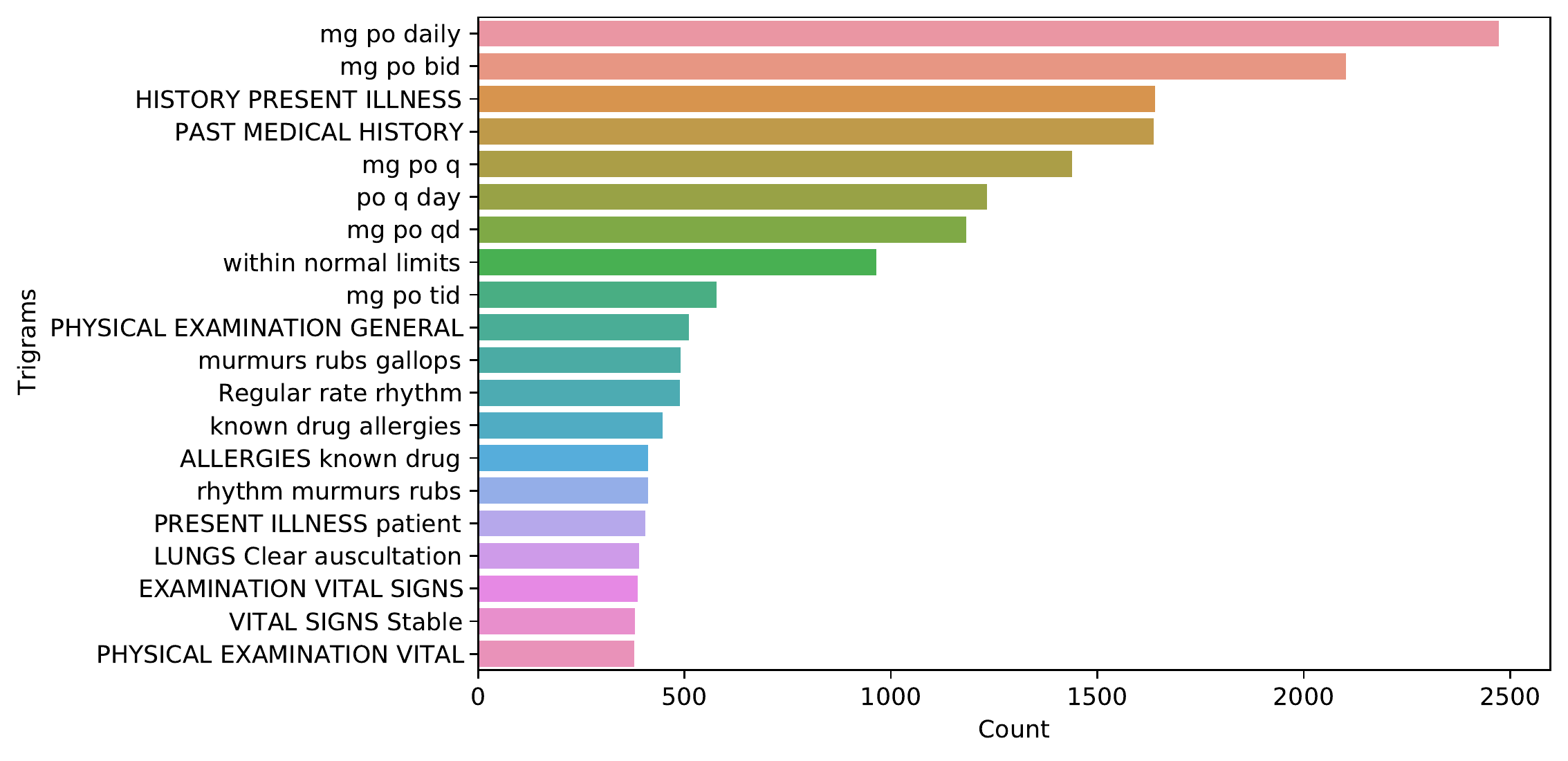}}
        \caption{Top 20 most frequent bi- and tri-grams in the UMN dataset (excluding any n-grams with digits)}
        \label{figure:umn_bi_tri_grams}
\end{figure}

Looking deeper into the \ABVS{}, there are on average 2.92 labels per \ABV{} and the dispersion between the minimum and maximum number of senses is 1 and 8, respectively.  
Figure~\ref{figure:umn_ambiguous_abv} displays the most ambiguous \ABVS{} in UMN along with each of their top labels. Compared to the MeDAL, the dataset is less imbalanced with a minimum of 5 and a maximum of 1,774 examples for a label, respectively. Finally, figures~\ref{fig:bi_UMN} and~\ref{fig:tri_UMN} display the top bi- and tri-grams in the UMN dataset. We note that the text is more unstructured than the MeDAL dataset and, as expected, the distribution is centered around prescriptions and patient analysis. 

\subsubsection{Data preprocessing}
In our numerical analysis, to focus the models on more topic-specific terms and also to reduce training time, punctuation and stop-words were removed from the text, any rows with \ABVS{} beyond the 110th word index were dropped, and text columns were truncated to a maximum length of 115 words. Additionally, to satisfy token classification model requirements, each token was mapped to its corresponding label; if the token was a regular word, it was assigned \textit{NA\_word}, and if it was an \ABV{}, it was mapped to its corresponding sense. 

The MeDAL dataset has an abundance of labels with an excessive amount of examples. While this does not necessarily hinder prediction performance, using all training examples can increase training time without significant performance improvements. Accordingly, for the MeDAL dataset, we dropped any instance where all labels in the row had at least 500 other rows to reference. These steps reduced the dataset by half from 147,728 rows to 73,196. 

Both datasets are further subsampled to ensure complete training, reduce the training time and avoid resource limitations in certain experiments. Therefore, only a subset of each dataset containing their respective top 12 most frequent \ABVS{} and 40 most frequent labels was used. After applying the same preprocessing steps as described for MeDAL, 8,472 rows remain in the UMN subset. 
We refer to the corresponding datasets as MeDAL-40 and UMN-40, respectively, in the rest of the paper.

\subsection{Experimental setup}
\begin{figure}[!ht]
        \centering
      
        \subfloat[Token classification pipeline\label{fig:token_diagram}]{ \includegraphics[width = 0.9\textwidth]{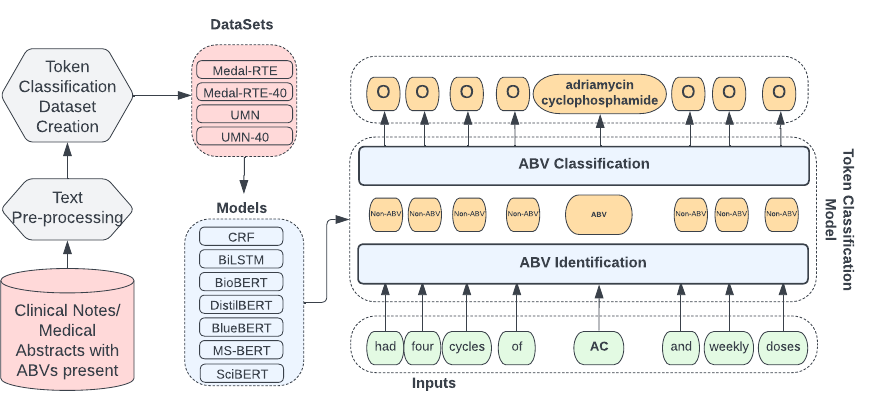}}
        \hfill
         \subfloat[Text classification pipeline with window of size 2 \label{fig:txt_diagram}]{\includegraphics[width=0.9\textwidth]{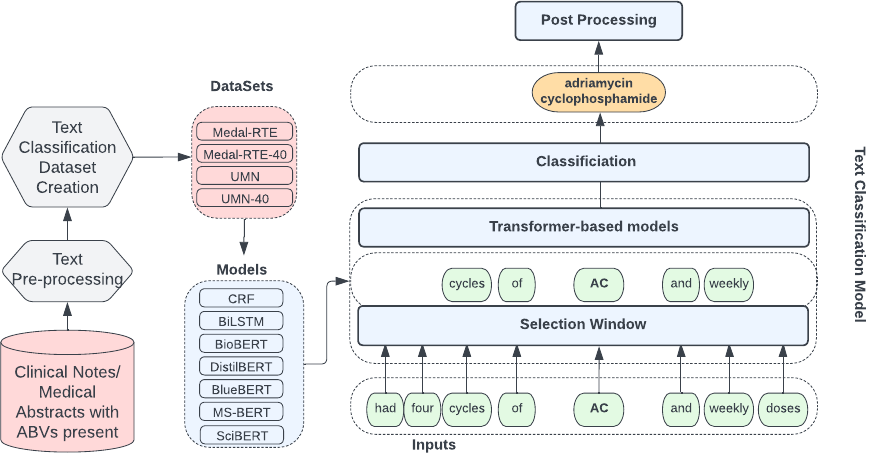}}
        \caption{Flowchart of token and text classification methods}
        \label{fig:ner_diagram}
\end{figure}

We evaluate our proposed token classification method for AD using different models over MeDAL and UMN datasets. Five different BERT-based models including DistilBERT, BioBERT, BlueBERT, MSBERT, and SciBERT are employed for the comparative analysis. We select BiLSTM and CRF as two baseline models to test our approach for the full-sized and limited 40-label versions of both datasets. Moreover, we compare the performance of our token classification approach with popular text classification methods for AD.
 Figure~\ref{fig:token_diagram} provides a visual depiction of the proposed token classification pipeline. The models receive the coded version of the dataset with \textit{NA\_word} and \ABV{} senses and classify both the common words and the target \ABV{}.

The standard text classification approach for AD is illustrated in Figure~\ref{fig:txt_diagram}. Text classification models select a window of $m$ words around the target \ABV{} and pass the data to Transformer-based models and the classifier. In this study, a window size of 40 is selected through extensive hyperparameter tuning experiments. Finally, a \textit{postprocessing} approach is implemented on the results of text classification models. In this approach, the predicted probability output is redistributed over the possible labels of the target \ABV{}, and the label with maximum probability is selected as the output.
Note that these possible labels are identified based on the corresponding labels of the \ABV{} in the training data. 
However, this method is not implemented on token classification raw outputs, since the methodology of the token classification incorporates the specific \ABVS{} for each sample, rendering the postprocessing redundant. Hyperparameter tuning experiments are employed for all the models and datasets and the final hyperparameters are reported in Table~\ref{table:model_hyper}.
\setlength{\tabcolsep}{4pt}
\renewcommand{\arraystretch}{1.15}
\begin{table}[!ht]
    \caption{Selected model hyperparameters for each dataset}
    \centering
    \resizebox{0.95\textwidth}{!}{
\begin{tabular}{l l P{130mm}} 
\toprule
 \textbf{Model} & \textbf{Dataset} & \multicolumn{1}{l}{\textbf{Hyperparameters}} \\ [0.5ex] 
\midrule
 \multirow{2}{*}{CRF} & MeDAL & algorithm : \textit{lbfgs}, c1 : 0.1, c2 : 0.1, max\_iterations : 100\\\cmidrule(lr){3-3}
  & UMN & algorithm : \textit{lbfgs}, $c1$ : 0.1, c2 : 0.1, max\_iterations : 100\\  
  \midrule
 \multirow{2}{*}{BiLSTM} & \multirow{1}{*}{MeDAL} & dropout\_rate : 0.3, optimizer : \textit{Adam}, learning\_rate : $5\text{e-}3$, activation : \textit{softmax}, loss\_function : \textit{sparse\_categorical\_crossentropy}, num\_epochs : 30, batch\_size : 64\\ \cmidrule(lr){3-3}

 & \multirow{1}{*}{UMN} & dropout\_rate : 0.3, optimizer : \textit{Adam}, learning\_rate : $5\text{e-}3$, activation : \textit{softmax}, loss\_function : \textit{sparse\_categorical\_crossentropy}, num\_epochs : 30, batch\_size : 64\\
  \midrule
 \multirow{1}{*}{DistilBERT} & \multirow{3}{*}{MeDAL} &  \multirow{3}{130mm}{max\_sequence\_length : 512, num\_epochs : 5, learning\_rate : $2\text{e-}5$, weight\_decay : 0.01, batch\_size = 8}\\
 \multirow{1}{*}{BioBERT} & \\
 \multirow{1}{*}{BlueBERT} & &  \\\cmidrule{3-3} 
 MS-BERT & \multirow{2}{*}{UMN} &  \multirow{2}{130mm}{max\_sequence\_length : 512, num\_epochs : 6, learning\_rate : $2\text{e-}5$, weight\_decay : 0.01, batch\_size = 8}\\ 
 \multirow{1}{*}{SciBERT} & &  \\
\bottomrule

\end{tabular}
}

 \label{table:model_hyper}
\end{table}

\begin{figure}[!ht]
            \centering
            \includegraphics[width = 0.3\textwidth]{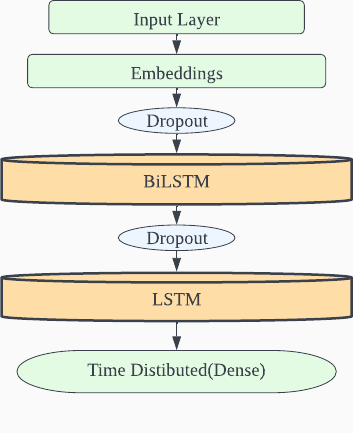}
            \caption{BiLSTM model architecture}
            \label{figure:lstm_model}
        \end{figure}

The employed BiLSTM structure in our analysis is presented in Figure~\ref{figure:lstm_model}. The model consists of an embedding layer, two dropout layers, two LSTM layers, one of which is Bidirectional, and a final TimeDistributed layer to output a label for every input token. BiLSTM uses the \textit{Adam} optimizer with a learning rate of 0.005, and \textit{sparse categorical cross-entropy} loss function to deal with the non-one-hot encoded labels.  Additionally, to decrease the impact of the imbalanced distribution of labels with the high majority of tokens being assigned to non-\ABV{} label (\textit{NA\_word}), we set the weight of all labels to 100 and kept \textit{NA\_word} at 1.

The training process for the pre-trained BERT models is as follows. We first tokenized the datasets using the respective BERT Fast Tokenizer and then created training batches using the \textit{DataCollatorForTokenClassification} class which also dynamically pads the sequences in each batch to be the same length. We then fine-tuned the model using the \textit{AutoModelForTokenClassification} class from HuggingFace\footnote{https://huggingface.co/}. 

The performance of the trained models is evaluated based on the 3-fold cross-validation method. 
Macro- and Weighted-F1 scores are the main performance metrics reported in this study. The Macro-F1 score calculation includes the \textit{NA\_word} label (the label that should be assigned to non-\ABV{} tokens), whereas, for Weighted-F1 score, it is excluded due to its abundantly large support and low importance. Macro- and Weighted-precision and recall values are also included to provide a more complete performance analysis and compare the models on a deeper level.

\section{Numerical results} \label{sec:Results}
In this section, we first compare the performance of different token classification methods for the AD task. Next, we select the best-performing model from the first part and compare its performance with different text classification methods. Finally, we explore the performances of the BERT-based token classification methods in comparison to the popular CRF approach over the limited label MeDAL-40 and UMN-40 datasets.

\subsection{Comparative analysis of token classification models}\label{sec:Comp_token}

\begin{figure}[!ht]
\centering
\subfloat[Macro average metrics for MeDAL dataset \label{fig:Medal_macro}]{\includegraphics[width=0.9\textwidth]{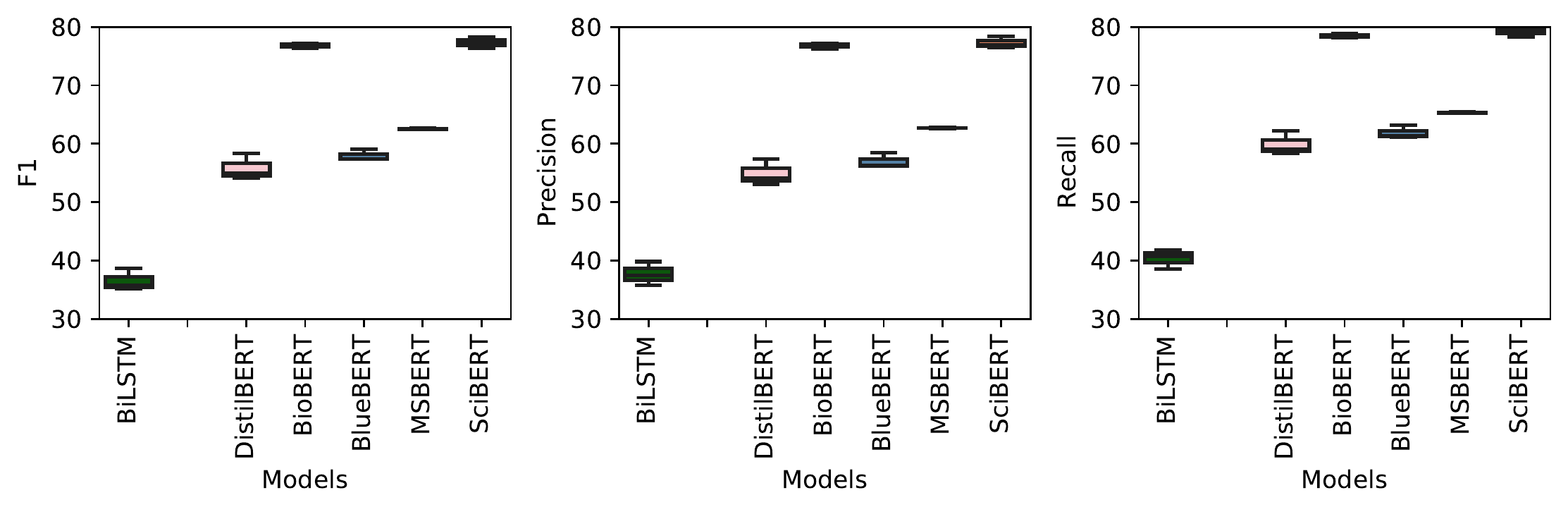}}
\hfill
\subfloat[Macro average metrics for UMN dataset \label{fig:UMN_macro}]{\includegraphics[width=0.9\textwidth]{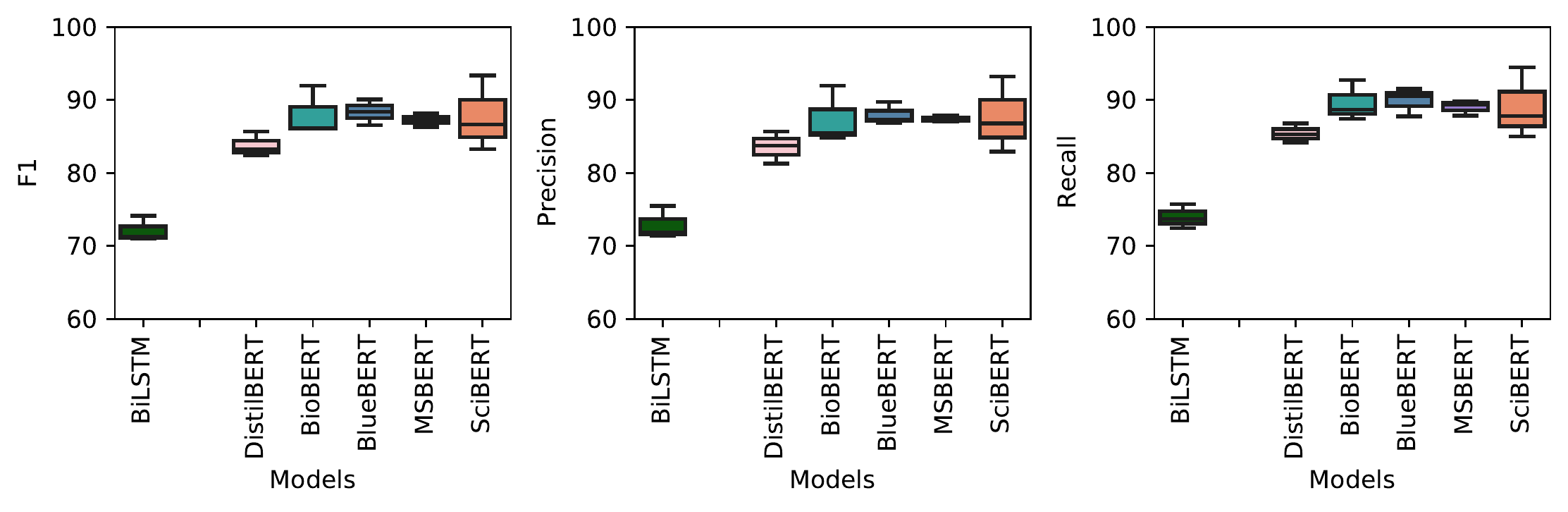}}

\caption{Token classification model performance comparison over full-label datasets.}
\label{fig:medal_UMN_token}
\end{figure}

In this experiment, we compare the performance of the BERT-based token classification methods against the BiLSTM model. Experiments are conducted over MeDAL and UMN datasets with more than 1,000 and 200 unique labels, respectively. 
We note that the large size of these datasets closely resembles the computational complexity of real-world scenarios. 
Table~\ref{table:fulllab_comp} presents the results of the experiments over both datasets and Figure~\ref{fig:medal_UMN_token} shows the box plot of Macro-F1 scores as obtained over 3 folds. 
\setlength{\tabcolsep}{6pt}
\renewcommand{\arraystretch}{1.5}
\begin{table}[!ht]
\caption{Summary performance values for the token classification models on full-label datasets. Results are averaged over 3 folds (highest scores on each metric are in bold).}
\centering
\resizebox{0.975\textwidth}{!}{
\begin{tabular}{c l r r r r r r r}
\toprule
\multirow {2}{*}{\textbf{Dataset}} & \multirow {2}{*}{\textbf{Model}} &   \multicolumn{3}{c}{\textbf{Macro Average}} &
\multicolumn{4}{c}{\textbf{Weighted Average}} \\
\cmidrule(lr){3-5}\cmidrule(lr){7-9}
 & &  F1(\%) & Precision(\%) & Recall(\%)& &  F1(\%) & Precision(\%) & Recall(\%) \\
\midrule
\multicolumn{1}{c}{\multirow{6}{*}{\STAB{\rotatebox[origin=c]{90}{{\textbf{MeDAL}}}}}}& \multicolumn{1}{l}{BiLSTM} & 36.57 $\pm$ 1.52&
37.71 $\pm$ 1.65&
40.34 $\pm$ 1.38&&
66.80 $\pm$ 1.96&
67.05 $\pm$ 1.56&
68.71 $\pm$ 2.01  \\
\cmidrule(lr){2-9}
 &\multicolumn{1}{l}{DistilBERT} & 55.80 $\pm$ 1.82&
54.88 $\pm$ 1.84&
59.88 $\pm$ 1.69&&
78.02 $\pm$ 1.69&
77.43 $\pm$ 1.97&
81.12 $\pm$ 1.24 \\
 &\multicolumn{1}{l}{BioBERT} &  76.79 $\pm$ 0.37 &76.81 $\pm$ 0.42& 78.48 $\pm$ 0.28 && 90.29 $\pm$ 0.68& 90.69 $\pm$ 0.89& 91.19 $\pm$ 0.37\\
 &\multicolumn{1}{l}{BlueBERT} & 57.94 $\pm$ 0.79&
56.95 $\pm$ 1.08&
61.87 $\pm$ 0.90&&
79.37 $\pm$ 0.86&
78.83 $\pm$ 1.59&
82.29 $\pm$ 0.41\\ 
 &\multicolumn{1}{l}{MS-BERT} & 62.56 $\pm$ 0.08&
62.69 $\pm$ 0.05&
65.32 $\pm$ 0.11&&
81.11 $\pm$ 0.92&
81.57 $\pm$ 1.29&
83.13 $\pm$ 0.67 \\ 
 &\multicolumn{1}{l}{SciBERT} & \bf 77.29 $\pm$ 0.74 &  \bf77.25 $\pm$ 0.82 &
 \bf79.16 $\pm$ 0.65 &&
 \bf90.53 $\pm$ 1.05 &
 \bf90.85 $\pm$ 1.37 &
 \bf91.58 $\pm$ 0.74\\ 
\toprule
\multicolumn{1}{c}{\multirow{6}{*}{\STAB{\rotatebox[origin=c]{90}{{\textbf{UMN}}}}}}
 &\multicolumn{1}{l}{BiLSTM} & 72.13 $\pm$ 1.42& 
72.90 $\pm$ 1.84& 
73.96 $\pm$ 1.36& & 
88.40 $\pm$ 0.29& 
88.10 $\pm$  0.23 &
89.60 $\pm$ 0.38\\ 
\cmidrule(lr){2-9}
 &\multicolumn{1}{l}{DistilBERT} & 83.75 $\pm$ 1.41&
83.57 $\pm$ 1.81&
85.41 $\pm$ 1.07&&
92.53 $\pm$ 0.96&
92.54 $\pm$ 1.46&
93.58 $\pm$ 0.53  \\
 &\multicolumn{1}{l}{BioBERT} &88.03 $\pm$ 2.75&
87.42 $\pm$ 3.20&
89.61 $\pm$ 2.26&&
 \bf95.05 $\pm$ 0.83&
94.79 $\pm$ 1.32&
 \bf95.97 $\pm$ 0.30 \\
 &\multicolumn{1}{l}{BlueBERT} & \bf 88.34 $\pm$ 1.42 &
 \bf87.98 $\pm$ 1.26&
 \bf89.92 $\pm$ 1.59 &&
94.32 $\pm$ 0.68&
94.30 $\pm$ 1.16&
95.18 $\pm$ 0.46\\
 &\multicolumn{1}{l}{MS-BERT} & 87.26 $\pm$ 0.77&
87.38 $\pm$ 0.30&
89.01 $\pm$ 0.85&&
93.87 $\pm$ 0.54&
94.47 $\pm$ 0.78&
94.71 $\pm$ 0.54\\
 &\multicolumn{1}{l}{SciBERT} & 87.73 $\pm$ 4.21&
87.64 $\pm$ 4.24&
89.09 $\pm$ 3.97&&
94.69 $\pm$ 1.14&
 \bf 94.92 $\pm$ 1.36&
95.29 $\pm$ 0.86 \\ \bottomrule
 
\end{tabular}
}
\label{table:fulllab_comp}
\end{table}

We observe that all the BERT-based models outperform the baseline BiLSTM model by a large margin. For the MeDAL dataset, SciBERT is the best performing model overall, with a Macro-F1 score of 77.29\%. On the other hand, the results show that all BERT models perform similarly for the UMN dataset. 
The significant difference between macro and weighted metrics highlights the imbalanced nature of the dataset. 
In particular, limited samples of specific labels deteriorate the macro metrics of the models.
BioBERT and SciBERT for the MeDAL and BlueBERT for the UMN are found to be the best-performing models. 
The strong performance by BioBERT is expected as it was pre-trained on text data from a similar domain. 
BlueBERT was pre-trained on similar text as well, though on a more limited basis, and hence achieved much lower macro scores for the MeDAL dataset. 
BlueBERT's limitations on MeDAL dataset could be explained by the class imbalance issue that affects its performance particularly for macro average metrics.

\subsection{Comparison against text classification methods}\label{sec:token_vs_text}
\begin{figure}[!ht]
\centering
\subfloat[Macro average metrics for MeDAL dataset \label{fig:Medal_macro_txt}]{\includegraphics[width=0.9\textwidth]{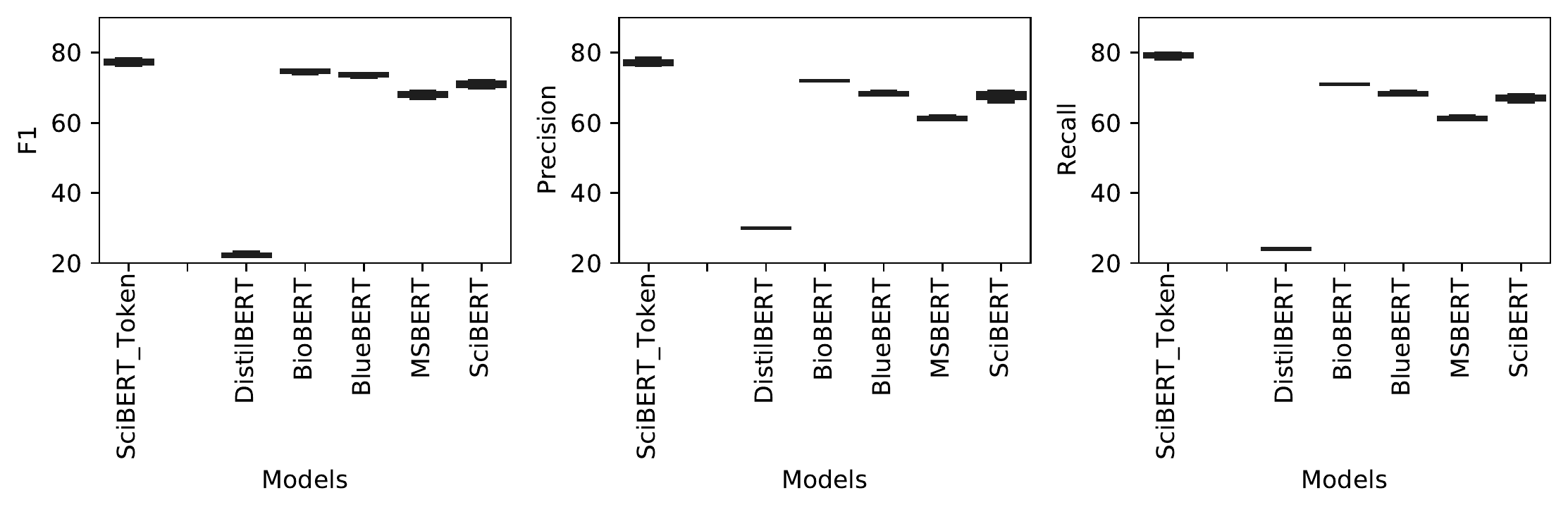}}
\hfill
\subfloat[Macro average metrics for UMN dataset \label{fig:UMN_macro_txt}]{\includegraphics[width=0.9\textwidth]{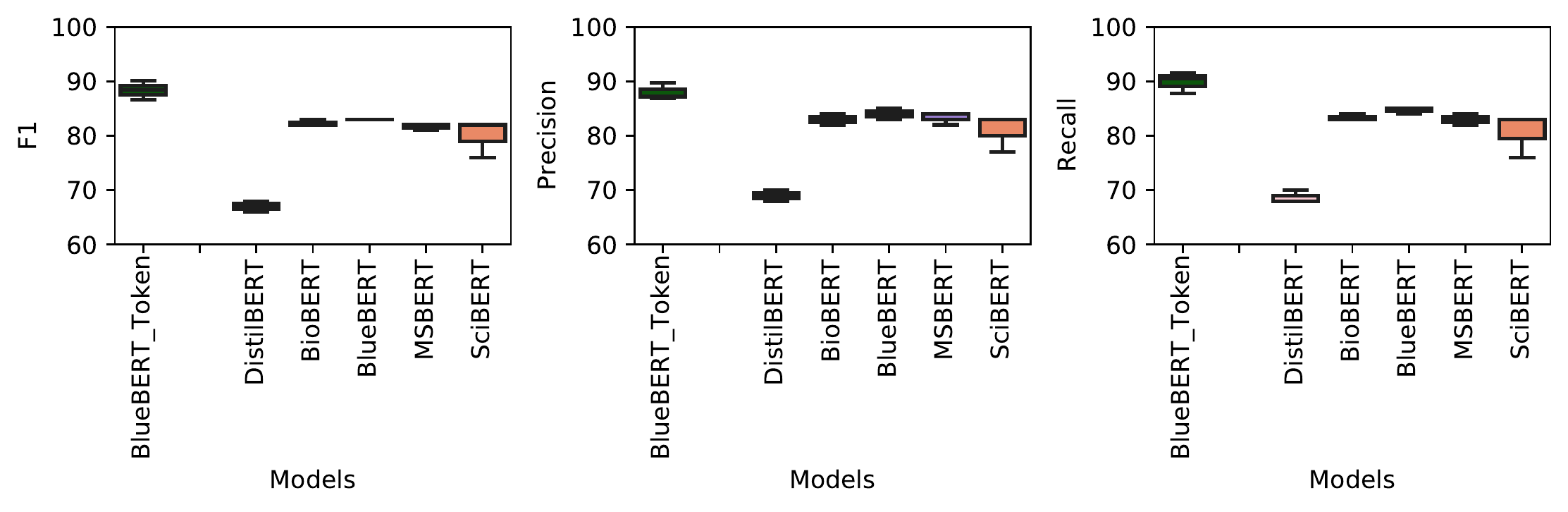}}

\caption{Text classification model performance comparison over full-label datasets.}
\label{fig:medal_UMN_txt}
\end{figure}

Table~\ref{table:textclass_comp} presents the postprocessed performances of different text classification models against the best-performing token classification model from the previous section. Moreover, the macro metrics are illustrated in Figure~\ref{fig:medal_UMN_txt}. The impact of postprocessing on the results of the text classification models is thoroughly explored and reported in Section~\ref{sec:secA1} of the Appendix.
\setlength{\tabcolsep}{6pt}
\renewcommand{\arraystretch}{1.5}
\begin{table}[!ht]
\caption{Summary performance values for the text classification models on full-label datasets and comparison against token classification baseline. Results are averaged over 3 folds (highest scores on each metric are in bold).}
\centering
\resizebox{0.975\textwidth}{!}{
\begin{tabular}{c l r r r r r r r}

\toprule
\multirow {2}{*}{\textbf{Dataset}} & \multirow {2}{*}{\textbf{Model}} &   \multicolumn{3}{c}{\textbf{Macro Average}} &
\multicolumn{4}{c}{\textbf{Weighted Average}} \\
\cmidrule(lr){3-5}\cmidrule(lr){7-9}
 & &  F1(\%) & Precision(\%) & Recall(\%)& &  F1(\%) & Precision(\%) & Recall(\%) \\
\midrule
\multicolumn{1}{c}{\multirow{9}{*}{\STAB{\rotatebox[origin=c]{90}{{\textbf{MeDAL}}}}}}& \multicolumn{7}{c}{\textit{Text Classification Models}}  \\ \cmidrule(lr){2-9}
&\multicolumn{1}{l}{DistilBERT}  & 22.33 $\pm$ 0.47&
30.0 $\pm$ 0.0&
24.0 $\pm$ 0.0&&
59.33 $\pm$ 0.47&
71.67 $\pm$ 0.47&
63.67 $\pm$ 0.47\\
 &\multicolumn{1}{l}{BioBERT} &74.67 $\pm$ 0.47&
72.0 $\pm$ 0.0&
71.0 $\pm$ 0.0&&
\bf 90.67 $\pm$ 0.47&
\bf 91.67 $\pm$ 0.47&
90.67 $\pm$ 0.47\\
 &\multicolumn{1}{l}{BlueBERT} &73.67 $\pm$ 0.47&
68.33 $\pm$ 0.47&
68.33 $\pm$ 0.47&&
88.0 $\pm$ 0.0&
89.0 $\pm$ 0.0&
88.0 $\pm$ 0.0\\
 &\multicolumn{1}{l}{MS-BERT}&68.0 $\pm$ 0.82&
61.33 $\pm$ 0.47&
61.33 $\pm$ 0.47&&
84.67 $\pm$ 0.47&
85.67 $\pm$ 0.47&
84.67 $\pm$ 0.47\\
& \multicolumn{1}{l}{SciBERT}  &71.0 $\pm$ 0.82&
67.67 $\pm$ 1.24&
67.0 $\pm$ 0.82&&
89.33 $\pm$ 0.47&
90.67 $\pm$ 0.47&
89.67 $\pm$ 0.47\\ \cmidrule(lr){2-9}
 & \multicolumn{7}{c}{\textit{Token Classification Baseline}} \\ \cmidrule(lr){2-9}
 &\multicolumn{1}{l}{SciBERT} & \bf 77.29 $\pm$ 0.74 &  \bf 77.25 $\pm$ 0.82 &
\bf 79.16 $\pm$ 0.65 &&
90.53 $\pm$ 1.05 &
90.85 $\pm$ 1.37 &
\bf 91.58 $\pm$ 0.74 \\ \toprule
\multicolumn{1}{c}{\multirow{9}{*}{\STAB{\rotatebox[origin=c]{90}{{\textbf{UMN}}}}}}& \multicolumn{7}{c}{\textit{Text Classification Models}}  \\ \cmidrule(lr){2-9}
&\multicolumn{1}{l}{DistilBERT}  & 67.0 $\pm$ 0.82 &
69.0 $\pm$ 0.82 &
68.67 $\pm$ 0.94 &
&
89.67 $\pm$ 0.47 &
90.0 $\pm$ 0.0 &
90.67 $\pm$ 0.47\\
 &\multicolumn{1}{l}{BioBERT} &82.33 $\pm$ 0.47 &
83.0 $\pm$ 0.82 &
83.33 $\pm$ 0.47 &
&
\bf 97.0 $\pm$ 0.0 &
\bf 97.0 $\pm$ 0.0 &
\bf 97.33 $\pm$ 0.47 \\
 &\multicolumn{1}{l}{BlueBERT} &83.0 $\pm$ 0.0 &
84.0 $\pm$ 0.82 &
84.67 $\pm$ 0.47 &
&
96.0 $\pm$ 0.0 &
96.0 $\pm$ 0.0 &
96.0 $\pm$ 0.0\\
 &\multicolumn{1}{l}{MS-BERT} & 81.67 $\pm$ 0.47 &
83.33 $\pm$ 0.94 &
83.0 $\pm$ 0.82 &
&
95.33 $\pm$ 0.47 &
96.0 $\pm$ 0.0 &
96.0 $\pm$ 0.0\\
& \multicolumn{1}{l}{SciBERT} &  80.0 $\pm$ 2.83 &
81.0 $\pm$ 2.83 &
80.67 $\pm$ 3.3 &
&
95.67 $\pm$ 1.89 &
95.67 $\pm$ 1.89 &
96.0 $\pm$ 1.41\\ 
\cmidrule(lr){2-9}
 & \multicolumn{7}{c}{\textit{Token Classification Baseline}} \\ \cmidrule(lr){2-9}
 &\multicolumn{1}{l}{BlueBERT} &  \bf 88.34 $\pm$ 1.42 &
 \bf 87.98 $\pm$ 1.26&
 \bf 89.92 $\pm$ 1.59 &&
94.32 $\pm$ 0.68&
94.30 $\pm$ 1.16&
95.18 $\pm$ 0.46\\ \bottomrule
 
\end{tabular}
}
\label{table:textclass_comp}
\end{table}

The results for the MeDAL dataset show that SciBERT token classification model outperforms all the text classification models over macro metrics. 
However, for the weighted metrics, SciBERT, BioBERT, and BlueBERT models lead to comparable performances. Similarly, for the UMN dataset, the BlueBERT token classification method outperforms all the other models in terms of macro metrics. On the other hand, BioBERT shows higher weighted metrics.

Overall, closely examining the relative performance on both datasets reveal that the best token classification model was able to outperform the majority of the text classification models. 
That is, that token classification can provide similar or better performance for medical AD tasks while also being able to manage text with multiple unique \ABVS{}. Moreover, the difference between macro and weighted metrics shows the impact of the class imbalance for both MeDAL and UMN datasets.

\subsection{Comparison against baseline CRF model}
\begin{figure}[!ht]
\centering
\subfloat[Macro average metrics for MeDAL-40 dataset \label{fig:Medal_40_macro}]{\includegraphics[width=0.9\textwidth]{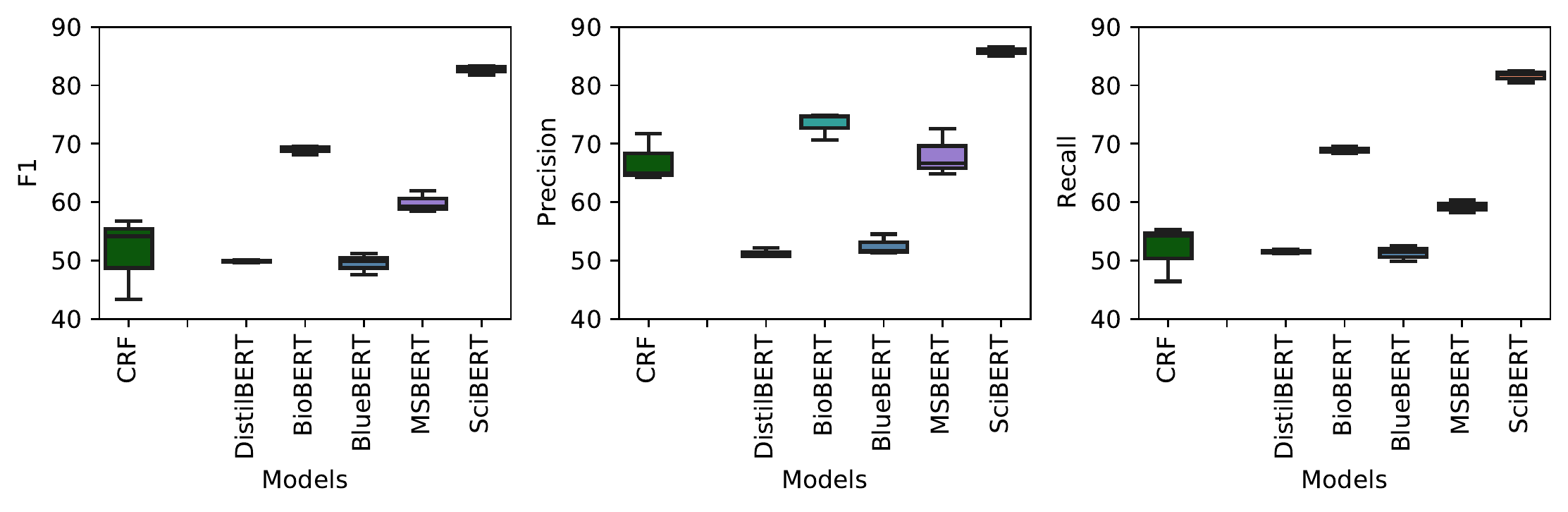}}
\hfill
\subfloat[Macro average metrics for UMN-40 dataset \label{fig:UMN_40_macro}]{\includegraphics[width=0.9\textwidth]{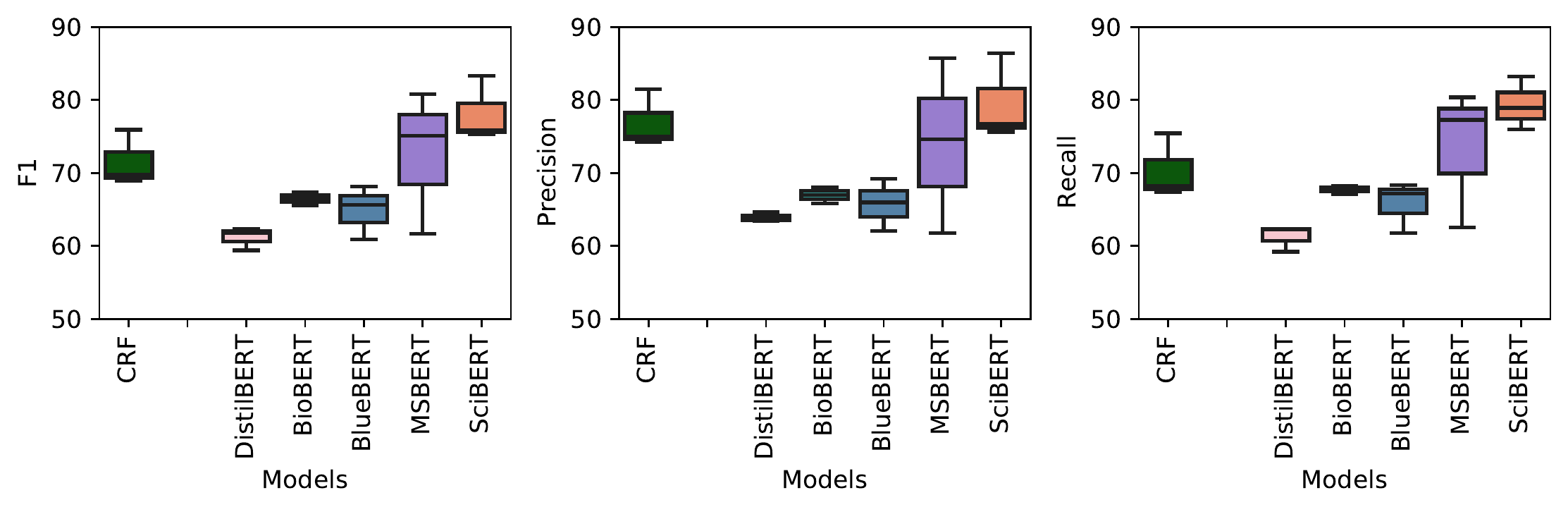}}

\caption{Token classification model performance comparison over limited-label datasets.}
\label{fig:medal_UMN_40_token}
\end{figure}

Table~\ref{table:40lab_comp} presents the performances of different token classification models against the CRF baseline model on a subsampled AD task with a total of 40 unique labels. Results of macro averaged metrics are also illustrated in Figure~\ref{fig:medal_UMN_40_token}.
\setlength{\tabcolsep}{7pt}
\renewcommand{\arraystretch}{1.5}
\begin{table}[!ht]
\caption{Summary performance values for the token classification models on 40-label datasets. Results are averaged over 3 folds (highest scores on each metric are in bold).}
\centering
\resizebox{0.975\textwidth}{!}{
\begin{tabular}{c l r r r r r r r}
\toprule
\multirow {2}{*}{\textbf{Dataset}} & \multirow {2}{*}{\textbf{Model}} &   \multicolumn{3}{c}{\textbf{Macro Average}} &
\multicolumn{4}{c}{\textbf{Weighted Average}} \\
\cmidrule(lr){3-5}\cmidrule(lr){7-9}
 & &  F1(\%) & Precision(\%) & Recall(\%)& &  F1(\%) & Precision(\%) & Recall(\%) \\
\midrule
\multicolumn{1}{c}{\multirow{6}{*}{\STAB{\rotatebox[origin=c]{90}{{\textbf{MeDAL-40}}}}}} 
&\multicolumn{1}{l}{CRF} &51.39 $\pm$ 5.81&
66.98 $\pm$ 3.37&
51.96 $\pm$ 3.93&&
66.02 $\pm$ 11.57&
77.16 $\pm$ 7.36&
67.01 $\pm$ 9.78  \\ 
 \cmidrule(lr){2-9}
 &\multicolumn{1}{l}{DistilBERT} & 49.88 $\pm$ 0.20 &51.24 $\pm$ 0.66 &51.54 $\pm$ 0.27 & &
71.84 $\pm$ 0.49 &71.59 $\pm$ 0.56 &74.66 $\pm$ 0.43\\
   &\multicolumn{1}{l}{BioBERT} &68.99 $\pm$ 0.62&
73.40 $\pm$ 1.92&68.93 $\pm$ 0.48&&84.79 $\pm$ 0.16&84.63 $\pm$ 0.17&85.76 $\pm$ 0.20\\
&\multicolumn{1}{l}{BlueBERT} &49.54 $\pm$ 1.46&
52.52 $\pm$ 1.42&51.27 $\pm$ 1.12&&71.71 $\pm$ 1.40&71.56 $\pm$ 1.61&75.11 $\pm$ 1.14\\
 &\multicolumn{1}{l}{MS-BERT} & 59.88 $\pm$ 1.51&
68.03 $\pm$ 3.28&59.27 $\pm$ 0.92&&76.39 $\pm$ 1.05&75.88 $\pm$ 1.12&78.54 $\pm$ 0.96\\
 &\multicolumn{1}{l}{SciBERT} & \bf 82.68 $\pm$ 0.66 & \bf 85.85  $\pm$ 0.61& \bf 81.61 $\pm$ 0.83&& \bf 89.06 $\pm$ 0.30&\bf 88.73 $\pm$ 0.42&\bf 89.85 $\pm$ 0.14\\
 \midrule
\multicolumn{1}{c}{\multirow{6}{*}{\STAB{\rotatebox[origin=c]{90}{{\textbf{UMN-40}}}}}}  
&\multicolumn{1}{l}{CRF} &71.54 $\pm$ 3.12&
76.90 $\pm$ 3.23&
70.33 $\pm$ 3.63&&
91.97 $\pm$ 0.34&
92.22 $\pm$ 0.26&
92.94 $\pm$ 0.31\\
 \cmidrule(lr){2-9}
 &\multicolumn{1}{l}{DistilBERT} &61.19 $\pm$ 1.28&
63.91 $\pm$ 0.54& 
61.26 $\pm$ 1.46& & 
92.02 $\pm$ 0.93& 
91.48 $\pm$ 0.84& 
93.48 $\pm$ 0.76\\
 &\multicolumn{1}{l}{BioBERT} & 66.50 $\pm$ 0.74&
66.96 $\pm$ 0.93&
67.69 $\pm$ 0.46 && 93.84 $\pm$ 0.64&
93.24 $\pm$ 0.56&
94.99 $\pm$ 0.60\\
 &\multicolumn{1}{l}{BlueBERT} & 64.87 $\pm$ 3.02&
65.75 $\pm$ 2.93&
65.75 $\pm$ 2.88&&
93.51 $\pm$ 0.44&
92.90 $\pm$ 0.34&
94.85 $\pm$ 0.35 \\
 &\multicolumn{1}{l}{MS-BERT} &72.53 $\pm$ 8.02&
74.03 $\pm$ 9.81&
73.40 $\pm$ 7.76&&
94.63 $\pm$ 0.50&
94.08 $\pm$ 0.68&
95.56 $\pm$ 0.29\\ 
 &\multicolumn{1}{l}{SciBERT} & \bf 78.13 $\pm$ 3.64& 
\bf79.57 $\pm$ 4.84& 
\bf79.35 $\pm$ 2.96& & 
\bf95.14 $\pm$ 0.53&  
\bf95.02$\pm$ 0.61& 
\bf95.72 $\pm$ 0.42\\ 
\bottomrule
 
\end{tabular}
}
\label{table:40lab_comp}
\end{table}

We observe that the majority of BERT-based models except DistilBERT outperform the CRF approach for both datasets. Moreover, SciBERT outperforms all the models for both datasets. Similar to full-size datasets, BioBERT and BlueBERT lead to a similar performance for the UMN dataset. Although the labels are sub-sampled and reduced to 40 most frequent labels, the visible difference between macro and weighted metrics shows that the label distributions in these datasets remain imbalanced. On the other hand, we note that reducing the number of labels in these datasets has led to significant reductions in the training times with the average training time of the models dropping from 40 minutes to 6 minutes and 16 minutes to 2 minutes for the MeDAL and UMN datasets, respectively.

\section{Discussion and conclusions} \label{sec:Conclusions}

In this study, we investigated the ability of transformers-based text and token classification models for the medical AD task. While the majority of recent studies explore text classification methods for the medical AD task, we adopted a token classification approach for this problem where each word (or \ABV{}) is assigned a label.


The results showed all BERT-based models outperform the BiLSTM and CRF baseline models.
Across our two main experiments, we found that BioBERT consistently performed well on both datasets, and was notably resistant to data imbalance. Furthermore, its stronger relative performance on both MeDAL dataset variations could be attributed to the fact that it was pre-trained on 18B words from medical abstracts. Similarly, MS-BERT was among the top performers on the UMN datasets which can also be credited to its relevant pre-training corpora. However, the most notable outcome from these experiments was found in SciBERT results. This model consistently performed the best or close to the best in both experiments and datasets while not being pre-trained on nearly as many medical abstract words as BioBERT, let alone any pre-training on clinical notes like MS-BERT. This success can be attributed to the use of a domain-specific vocabulary, instead of the one created by BERT\textsubscript{base}, playing a key role in the performance of SciBERT, and should be a consideration in future works.
Overall, through our detailed experiments, we were able to examine the efficacy of token classification methods on the medical AD task and demonstrate its benefits over text classification-based approaches. 
Additionally, our results showed the effect of transfer learning, the significance of relevant pre-training corpora, and the importance of a model's resistance to label imbalance.


The datasets used in our study, while relevant, do not necessarily capture all possible environments where medical \ABVS{} could occur.
Accordingly, repeating our experiments with more datasets, ideally, clinical patient notes with multiple unique target \ABVS{} could be helpful in medical AD research, however, we note that suitable public medical datasets are rarely made available. Regarding our chosen solution to this task, using token classification methods on this problem only enhances the class imbalance issue due to the amount of non-\ABV{} entities present in each text. Moreover, limited hyperparameter tuning experiments can pose a threat to validity. Conducting more extensive hyperparameter tuning experiments could help further improve the performance for our medical AD tasks.

In our numerical analysis, we incorporated limited features for the CRF model, which may deteriorate the models' performance. 
Future work can benefit from a more extensive feature selection procedure.
Furthermore, the postprocessing was found to have significant impact for the text classification model performance. 
However, exploring and comparing alternative postprocessing methods could prove to be beneficial for this task. 
Finally, although the motivation of our study was to use token classification methods for the medical AD task, based on the results from the text classification baseline, another possible route could be to combine text classification and token classification methods into one comprehensive model, e.g., using a voting ensemble.


\section*{Declarations}
\subsection*{Ethical Approval}
Not applicable.

\subsection*{Competing interests}
Not applicable.

\subsection*{Authors contributions}
All the co-authors contributed to the conception, design, implementation, writing and review of the paper. Author order is alphabetical.

\subsection*{Funding}
Not applicable.

\subsection*{Availability of data and materials}
All the datasets used in our analysis are publicly available and the links to these datasets are provided as follows:
\begin{itemize}
    \item MeDAL: \href{https://www.kaggle.com/datasets/xhlulu/medal-emnlp}{https://www.kaggle.com/datasets/xhlulu/medal-emnlp}
    \item UMN: \href{https://conservancy.umn.edu/handle/11299/137703}{https://conservancy.umn.edu/handle/11299/137703}
\end{itemize}

\begin{appendices}

\section{Text classification postprocessing results }\label{sec:secA1}
In this section, we have reported the detailed results for the text classification experiments in Section~\ref{sec:token_vs_text}. Table~\ref{table:textclass_comp_post_pre} presents the performance values before and after applying postprocessing.
Overall, we observe that almost all the models benefit from postprocessing. In particular, DistilBERT, BlueBERT, and MS-BERT experiences a significant performance improvement for the MeDAL dataset. On the other hand, BioBERT and SciBERT models' performances do not benefit from the postprocessing approach on the UMN dataset.

\setlength{\tabcolsep}{2.0pt}
\renewcommand{\arraystretch}{1.7}
\begin{table}[!ht]
\caption{Summary performance values for Transformer-based text classification models for the full-label datasets with and without postprocessing. Results are averaged over 3 folds (highest scores on each metric are in bold).}
\centering
\resizebox{1.05\textwidth}{!}{
\-\hspace{-0.4cm}\begin{tabular}{c l c c c c c c c c c c c c c c c c}

\toprule
\multirow {3}{*}{\textbf{Dataset}} & \multirow {3}{*}{\textbf{Model}} & \multicolumn{7}{c}{\textbf{Raw results}} & & \multicolumn{7}{c}{ \textbf{Post-processed results}}\\
\cmidrule(lr){3-9} \cmidrule(lr){10-17}
&&\multicolumn{3}{c}{Macro Average} &&  
\multicolumn{3}{c}{Weighted Average} & & \multicolumn{3}{c}{Macro Average}
&& \multicolumn{3}{c}{Weighted Average} \\
\cmidrule(lr){3-5}\cmidrule(lr){7-9} \cmidrule(lr){11-13}\cmidrule(lr){15-17}
 & &  F1(\%) & Prec(\%) & Rec(\%)& &  F1(\%) & Prec(\%) & Rec(\%) & & F1(\%) & Prec(\%) & Rec(\%)& &  F1(\%) & Prec(\%) & Rec(\%)\\
\midrule
\multicolumn{1}{c}{\multirow{5}{*}{\STAB{\rotatebox[origin=c]{90}{{\textbf{MeDAL}}}}}}&\multicolumn{1}{l}{DistilBERT}  &0.0 $\pm$ 0.0 &
0.0 $\pm$ 0.0&
0.0 $\pm$ 0.0&&
0.0 $\pm$ 0.0&
0.0 $\pm$ 0.0&
0.0 $\pm$ 0.0 && 22.3 $\pm$ 0.5&
30.0 $\pm$ 0.0&
24.0 $\pm$ 0.0&&
59.3 $\pm$ 0.5&
71.7 $\pm$ 0.5&
63.7 $\pm$ 0.5\\
 &\multicolumn{1}{l}{BioBERT} &\bf 74.7 $\pm$ 0.5&
\bf72.0 $\pm$ 0.0&
\bf71.0 $\pm$ 0.0&&
\bf90.7 $\pm$ 0.5&
\bf91.7$\pm$ 0.5&
\bf90.7 $\pm$ 0.5 &&\bf74.7 $\pm$ 0.5&
\bf72.0 $\pm$ 0.0&
\bf71.0 $\pm$ 0.0&&
\bf90.7 $\pm$ 0.5&
\bf91.7 $\pm$ 0.5&
\bf90.7 $\pm$ 0.5\\
 &\multicolumn{1}{l}{BlueBERT} &12.3 $\pm$ 0.5&
11.0 $\pm$ 0.0&
10.0 $\pm$ 0.0&&
24.7 $\pm$ 0.5&
29.0 $\pm$ 0.0&
24.0 $\pm$ 0.0&&73.7 $\pm$ 0.5&
68.3 $\pm$ 0.5&
68.3 $\pm$ 0.5&&
88.0 $\pm$ 0.0&
89.0 $\pm$ 0.0&
88.0 $\pm$ 0.0\\
 &\multicolumn{1}{l}{MS-BERT} & 4.3 $\pm$ 0.5&
5.0 $\pm$ 0.0&
4.0 $\pm$ 0.0&&
11.0 $\pm$ 0.0&
17.0 $\pm$ 0.0&
12.0 $\pm$ 0.0&&68.0 $\pm$ 0.8&
61.3 $\pm$ 0.5&
61.3 $\pm$ 0.5&&
84.7 $\pm$ 0.5&
85.7 $\pm$ 0.5&
84.7 $\pm$ 0.5\\
& \multicolumn{1}{l}{SciBERT}  &57.0 $\pm$ 0.8&
53.3 $\pm$ 1.3&
53.0 $\pm$ 0.8&&
78.7 $\pm$ 0.5&
80.7 $\pm$ 0.5&
78.7 $\pm$ 0.5 & &71.0 $\pm$ 0.8&
67.7 $\pm$ 1.2&
67.0 $\pm$ 0.8&&
89.3 $\pm$ 0.5&
90.7 $\pm$ 0.5&
89.7 $\pm$ 0.5\\ 
\toprule
\multicolumn{1}{c}{\multirow{5}{*}{\STAB{\rotatebox[origin=c]{90}{{\textbf{UMN}}}}}}
&\multicolumn{1}{l}{DistilBERT}  & 14.3 $\pm$ 0.5 &
14.7 $\pm$ 0.9 &
16.7 $\pm$ 0.5 &
&
32.0 $\pm$ 0.8 &
31.3 $\pm$ 1.3 &
38.0 $\pm$ 0.8 &&
67.0 $\pm$ 0.8 &
69.0 $\pm$ 0.8 &
68.7 $\pm$ 0.94 &
&
89.7 $\pm$ 0.5 &
90.0 $\pm$ 0.0 &
90.7 $\pm$ 0.5 \\
 &\multicolumn{1}{l}{BioBERT} & \bf 82.3 $\pm$ 0.5 &
\bf83.0 $\pm$ 0.8 &
\bf83.3 $\pm$ 0.5 &
&
\bf97.0 $\pm$ 0.0 &
\bf97.0 $\pm$ 0.0 &
\bf97.3 $\pm$ 0.5 &&
82.3 $\pm$ 0.5 &
83.0 $\pm$ 0.8 &
83.3 $\pm$ 0.5 &
&
\bf97.0 $\pm$ 0.0 &
\bf97.0 $\pm$ 0.0 &
\bf97.3 $\pm$ 0.5 \\
 &\multicolumn{1}{l}{BlueBERT} &44.0 $\pm$ 0.0 &
47.3 $\pm$ 1.7 &
45.0 $\pm$ 0.0 &
&
67.7 $\pm$ 0.5 &
68.0 $\pm$ 0.8 &
70.0 $\pm$ 0.0 &&
\bf83.0 $\pm$ 0.0 &
\bf84.0 $\pm$ 0.8 &
\bf84.7 $\pm$ 0.5 &
&
96.0 $\pm$ 0.0 &
96.0 $\pm$ 0.0 &
96.0 $\pm$ 0.0 \\
 &\multicolumn{1}{l}{MS-BERT} & 37.3 $\pm$ 0.5 &
40.3 $\pm$ 0.5 &
38.3 $\pm$ 0.5 &
&
60.7 $\pm$ 0.5 &
61.0 $\pm$ 0.0 &
63.0 $\pm$ 0.0 &&
81.7 $\pm$ 0.5 &
83.3 $\pm$ 0.9 &
83.0 $\pm$ 0.8 &
&
95.3 $\pm$ 0.5 &
96.0 $\pm$ 0.0 &
96.0 $\pm$ 0.0 \\
& \multicolumn{1}{l}{SciBERT} &79.0 $\pm$ 2.8 &
80.3 $\pm$ 2.6 &
79.0 $\pm$ 2.8 &
&
94.3 $\pm$ 1.9 &
94.3 $\pm$ 1.9 &
94.7 $\pm$ 1.7 &&
80.0 $\pm$ 2.8 &
81.0 $\pm$ 2.8 &
80.7 $\pm$ 3.3 &
&
95.7 $\pm$ 1.9 &
95.7 $\pm$ 1.9 &
96.0 $\pm$ 1.4 \\ 
\bottomrule
 
\end{tabular}
}
\label{table:textclass_comp_post_pre}
\end{table}

\end{appendices}

\bibliographystyle{elsarticle-num-names}
\bibliography{refs}

\end{document}